%% file: main.tex
\documentclass[letterpaper, 10 pt, conference]{ieeeconf}

\IEEEoverridecommandlockouts
\usepackage{cite}
\usepackage{amsmath,amssymb,amsfonts}
\usepackage{algorithmic}
\usepackage[short]{optidef}
\usepackage{physics} % for \norm
\usepackage{graphicx}
\usepackage{textcomp}
\usepackage{xcolor}
\usepackage{hyperref}

\usepackage{bm}

\usepackage{tikz}
\usepackage[utf8]{inputenc}
\usepackage{pgfplots} 
\usepackage{pgfgantt}
\usepackage{pdflscape}
\pgfplotsset{compat=newest} 
\pgfplotsset{plot coordinates/math parser=false}
\def\*#1{\bm{#1}}

\begin{document}

\title{\LARGE \bf Flying Vines: Design, Modeling, and Control\\of a Soft Aerial Robotic Arm}
\author{Rianna Jitosho, Crystal E. Winston, Shengan Yang, Jinxin Li, Maxwell Ahlquist, Nicholas John Woehrle, \\C. Karen Liu, and Allison M. Okamura% <-this % stops a space
\thanks{This work was supported in part by the Achievement Rewards for College Scientists (ARCS) and National Science Foundation (NSF) grant 2345769.}% <-this % stops a space
\thanks{C. Karen Liu is with the Dept. of Computer Science, and the other authors are with the Dept. of Mechanical Engineering, Stanford University, Stanford, CA 94305, USA. Email: karenliu@cs.stanford.edu, \{rjitosho, cwinston, syang9, lijinxin, ahlquist, nickjw, aokamura\}@stanford.edu}%
}
% \author{Authors hidden for double blind review}
% \author{\IEEEauthorblockN{Rianna Jitosho}
% \and
% \IEEEauthorblockN{Crystal Winston}
% \and
% \IEEEauthorblockN{Jinxin Li}
% \and
% \IEEEauthorblockN{Austin Yang}
% \and
% \IEEEauthorblockN{Allison Okamura}
% \and
% \IEEEauthorblockN{C. Karen Liu}

\maketitle
\thispagestyle{empty}
\pagestyle{empty}

\begin{abstract}
Aerial robotic arms aim to enable inspection and environment interaction in otherwise hard-to-reach areas from the air. However, many aerial manipulators feature bulky or heavy robot manipulators mounted to large, high-payload aerial vehicles. Instead, we propose an aerial robotic arm with low mass and a small stowed configuration called a “flying vine”. The flying vine consists of a small, maneuverable quadrotor equipped with a soft, growing, inflated beam as the arm. This soft robot arm is underactuated, and positioning of the end effector is achieved by controlling the coupled quadrotor-vine dynamics. In this work, we present the flying vine design and a modeling and control framework for tracking desired end effector trajectories. The dynamic model leverages data-driven modeling methods and introduces bilinear interpolation to account for time-varying dynamic parameters. We use trajectory optimization to plan quadrotor controls that produce desired end effector motions. Experimental results on a physical prototype demonstrate that our framework enables the flying vine to perform high-speed end effector tracking, laying a foundation for performing dynamic maneuvers with soft aerial manipulators.
\end{abstract}

% \begin{IEEEkeywords}
% Modeling, Control, and Learning for Soft Robots, Aerial Vehicles, Underactuated Robots
% \end{IEEEkeywords}
% mechanics and control
% Flexible Robotics, Dynamics

\section{Introduction}

Aerial vehicles are well-suited for accessing hard-to-reach areas from the air, and augmenting these vehicles with robotic arms can broaden the areas they can access for inspection and enable new types of environment interaction. A straightforward approach to realizing aerial robotic arms is to mount traditional robot arms onto aerial vehicles, such as a serial or delta manipulator~\cite{Bouzgou2019, Cocuzza2020,Bodie2021}. Higher dexterity can be achieved by using arms with more degrees of freedom \cite{Sheng2020} or multiple arms attached to the aerial vehicle \cite{Paul2021}. While these solutions enable precise end effector positioning, they are often heavy and bulky, reducing flight time and payload capacity. Although weight can be reduced with lighter materials \cite{Perez-Jimenez2020, Sanchez-Cuevas2020}, there is still a steep trade-off between the arm workspace and the weight/size of the system. 

As an alternative to an active robot arm, suspended payloads via cable-based solutions are passive, lightweight, and small. This simplifies the hardware, enables carrying heavier payloads, and reduces the dynamic coupling of the ``arm" with the aerial vehicle \cite{autotrans}. Many works have explored how to model and control these systems \cite{Alothman2016, Cruz2017, Guo2020}, but cable-based solutions are inherently limited to slung-load applications.

Hardware solutions have emerged that aim to address different shortcomings of traditional aerial robotic arms and cable-based solutions. Collapsible designs aim to reduce the stowed size of the robot arm, which would reduce drag during flight, and prior work has leveraged scissor mechanisms~\cite{Suthar2021, Choi2022} as well as origami-inspired designs~\cite{Kim2018}. While these designs are compact when stowed and can have a long reach when deployed, they only reach directly below the aerial vehicle. Continuum arms have also been proposed because they offer the dexterity and flexibility of cables with more options for controlling the behavior of the system. Some prior works focus on modeling and control in simulation~\cite{Samadikhoshkho20, Szasz2022}, while others also present a physical prototype~\cite{Chien2023, peng2025dexterous}. However, these continuum arm designs have an inherent trade-off between workspace and stowed size.
% ============================
\begin{figure}
    \vspace{2mm}
    \centering
    \includegraphics[width=\linewidth]{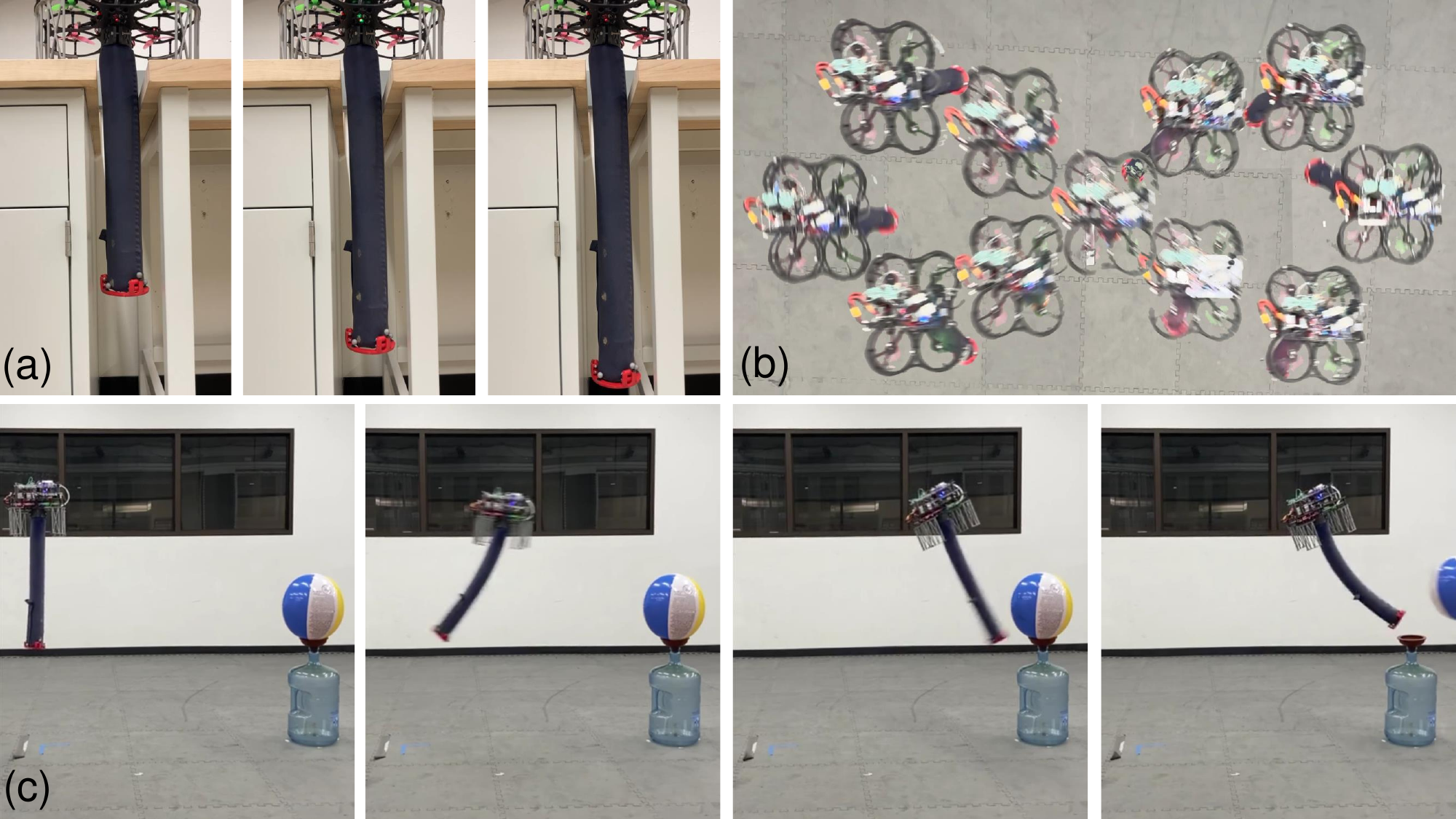}
    \vspace{-7mm}
    \caption{Images of the flying vine prototype in action. (a) Images from a video of the soft robot arm growing. (b) Composite image that overlays frames of a video in which the flying vine tracks a lemniscate path ($\infty$). (c) Images from a video of the flying vine ``kicking" a beach ball. }
    \label{fig:summary}
    \vspace{-4.5mm}
\end{figure}
% ============================
As in the prior works described above, we aim to address the shortcomings of traditional aerial robotic arms and cable-based solutions. Our choice of robot arm is a soft growing ``vine robot"; Fig.~\ref{fig:summary} and the supplementary video show the ``flying vine" prototype in action. The primary component of a vine robot is an inflated beam that is highly compact when stowed and can achieve significant length change as it ``grows" via eversion at the tip\cite{HawkesScienceRobotics2017}. This addresses the challenge faced by existing aerial robotic arms that must trade off workspace with weight and size. Additionally, while cable-based solutions are restricted to slung-load applications, a vine robot can vary its stiffness by adjusting its internal pressure, offering greater versatility. Existing work with vine robots has focused on ground-based \cite{CoadRAL2020} or ceiling-mounted~\cite{StroppaICRA2020} setups, but the vine robot implementation in this work is tailored for mounting onto a small quadrotor. A vine robot has previously been mounted on a quadrotor~\cite{phlosar}, but in this work, the integrated vine robot can change its length and internal pressure in real time, and we provide a modeling and control framework to control the position of the end effector.

The primary contribution of this work is demonstrating the feasibility and promise of flying vines for aerial manipulation using critical design insights paired with a modeling and control framework. We envision that the flying vine could be useful in aerial applications where a long, flexible robot arm is advantageous. For example, in bridge inspection~\cite{bridge_inspection}, a flying vine could deliver an end effector camera deep inside a pipe on a bridge. In canopy sampling~\cite{canopy_sampling}, a flying vine could deposit a canopy sensor while allowing the quadrotor to maintain a safe distance from tree branches. In this work, we do not tackle application-specific demonstrations and instead focus on foundational modeling and control tools that will enable these types of manipulation tasks.

% -------------------------------------------------------------
\section{Design}

\begin{figure}
    \vspace{2.5mm}
    \centering
    \includegraphics[width=\linewidth]{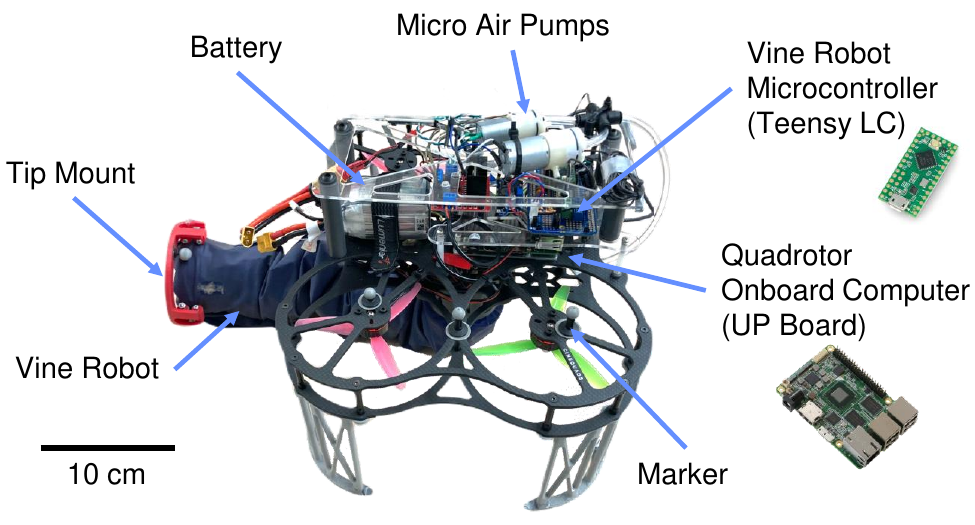}
    \vspace{-7mm}
    \caption{Close-up of the flying vine prototype. Micro air pumps are used to inflate the vine robot. The quadrotor has an onboard computer (UP Board) for processing commands and a microcontroller (Teensy LC) for executing vine robot commands. We mount reflective markers on the frame of the quadrotor and on the vine robot end effector, also referred to as the ``tip mount," for use with a motion capture system (OptiTrack).}
    \label{fig:prototype}
    \vspace{-4.5mm}
\end{figure}

At a high level, the flying vine is a quadrotor with a soft, growing, vine robot arm mounted to its underside (Fig.~\ref{fig:prototype}). The vine robot in this work (Fig.~\ref{fig:tip_spool}) is tailored for use on a small quadrotor, which requires a design that reduces weight and size while operating untethered.
% ============================
% ============================
\subsection{Vine Robot}

The primary component of a vine robot is an inflated beam that can grow via eversion. These beams are commonly formed by making tubes from plastic film or fabric. The flying vine uses 30-denier silicone impregnated nylon ripstop (Seattle Fabrics) since it is lightweight, flexible, and thin. Two micro air pumps (Amazon \#B078H8V563) are connected in parallel to inflate the vine robot, enabling untethered pneumatic operation. We attach the vine robot to the quadrotor with a clamp interface, where the clamp is mounted to the underside of the quadrotor, and it clamps onto a rigid hollow cylinder placed inside the vine robot.

% ============================
\begin{figure}
\vspace{2.5mm}
    \centering
    \includegraphics[width=\linewidth, trim=3mm 0 3in 0, clip,]{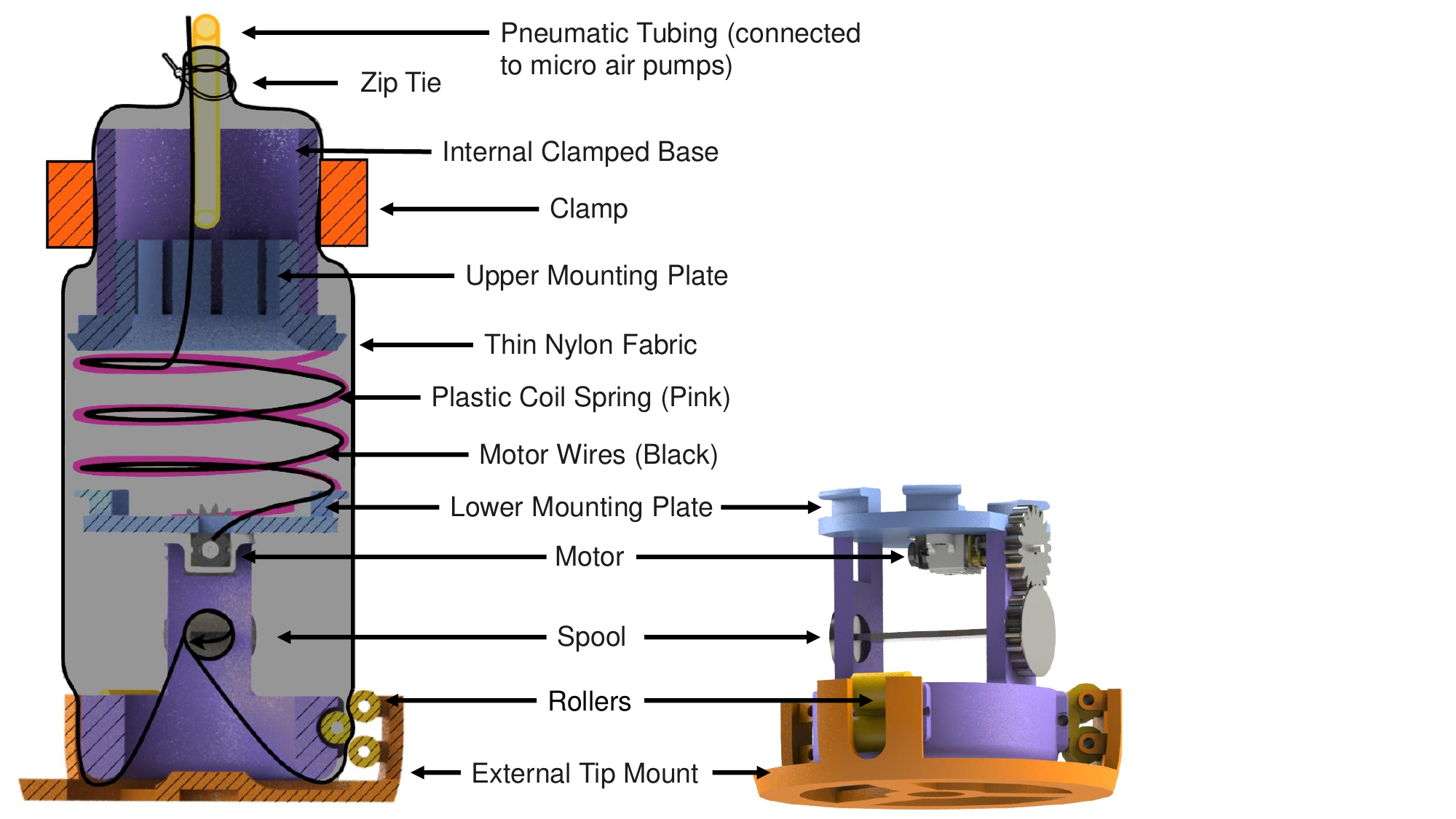}
    \vspace{-7mm}
    \caption{Design details of the soft growing vine robot arm. The vine robot is fixed to the quadrotor via a clamp mounted to the underside of the quadrotor. The vine ``grows" by rotation of the spool, which releases additional material that everts due to internal vine body pressure and gravity. Motor wires are routed along a plastic coil for passive wire management. The ends of the plastic coil are fixed to mounting plates, and the upper mounting plate is press-fit into the internal clamped base. Interlocking rollers keep the relative positioning between the internal spool and external tip mount fixed. The tip mount provides a rigid surface for mounting motion capture markers.}
    \label{fig:tip_spool}
    \vspace{-4.5mm}
\end{figure}
% ============================

Uneverted vine material is commonly stored on a spool at the proximal base of a vine robot. The spool and housing are often much larger than the vine robot diameter because the long length of spooled material enables the vine robot to extend many meters~\cite{blumenschein2020}. We trade off material storage space for a highly compact, lightweight spool adapted from the design of Haggerty et al.~\cite{srm}. The flying vine spool weighs about 100~g and is positioned inside the vine body (8~cm diameter) at its distal tip. Rotation of the spool is actuated by a 1000:1 Micro Metal Gearmotor (Pololu \#3070), and a combination of gravity and vine robot pressurization causes the unspooled material to evert, thus lengthening the vine robot. The current internal tip spool accommodates about 0.5~m of material due to the proximity of the motor to the spool, but this length could be increased by switching from a gear drive to a belt drive.

To simplify the mechatronic design by eliminating the need for wireless communication to the tip spool, we have a direct wire connection to the motor from the motor driver mounted on the quadrotor. Our approach for simple, passive cable management is to route the motor wires along a plastic coil spring (Amazon \#B08MBB4QKK), where one end of the coil is fixed to the proximal base of the vine robot and the other is fixed to the distal internal tip spool. This prevents the wires from tangling or interfering with the moving parts of the internal spool as the vine undergoes length change.

Commands are sent to the actuators through a microcontroller (Teensy LC). Both the motor and micro air pumps operate at 6V DC, and we use a dual H-bridge motor driver for these actuators (Amazon \#B0CLYBPGP9). When the microcontroller receives a growth rate command, it sends a pulse width modulation (PWM) command to the motor driver. The growth rate command corresponds to the motor direction and PWM duty cycle. The microcontroller also runs an on-off controller to operate the micro air pumps, and when the microcontroller receives a pressure command, it updates the on-off controller setpoint. The maximum vine growth rate for the physical prototype is approximately 2~cm/s, and the maximum flow rate of a single micro air pump is approximately 3.2~L/min. 

Experiments are performed in a room equipped with an OptiTrack motion capture system. Offboard OptiTrack cameras perceive passive reflective markers attached to the object of interest (Fig.~\ref{fig:prototype}). We require end effector position measurements for the modeling and control framework, which in turn requires placing reflective markers at the tip of the vine robot. The simplest option would be to attach the markers to the vine robot fabric, however, as the vine grows, newly everted fabric would be more distal than the existing reflective markers. Instead, we draw from existing vine robot literature and utilize a ``tip mount"~\cite{roboa}. As illustrated in Fig.~\ref{fig:tip_spool}, the internal tip spool and the external tip mount feature interlocking rollers through which the vine robot fabric routes. This ensures that the external tip mount moves in tandem with the internal tip spool without interfering with material eversion. The external tip mount serves as the end effector of the flying vine. It could be adapted to have features for environment interaction (camera, passive hook, active gripper, etc), although in this work we only use it to mount motion capture markers.

\subsection{Aerial Vehicle}

The flying vine quadrotor has 12.7 cm propellers and 2300 kv brushless motors (RaceSpec RS2205) mounted on a Lumenier QAV-PRO frame and is powered with a 14.8V Lipo battery (Tattu R-Line). The maximum thrust of the quadrotor is 4000 gf, and the vine robot, along with its electronics and mounting hardware, takes up less than 650~g. The complete flying vine weighs about 1700~g and is roughly 29x34x26~cm in its stowed configuration.
% frame 3.5cm top 8cm bottom 13 cm

The quadrotor flight controller (PixHawk Pro) has an internal controller to track a position and yaw command. For simplicity, we only use the position command and leave the yaw command at zero. While it is possible to bypass the position command and send lower-level control commands such as velocity commands, thrust and torque commands, or even motor voltage commands, we chose to keep the position command infrastructure to leverage pre-tuned parameters and preserve the built-in safety features for better vehicle stability.

\subsection{Sensing and Communication}
\label{sec:comms}

\begin{figure}
\vspace{2.5mm}
    \centering
    \includegraphics[width=\linewidth]{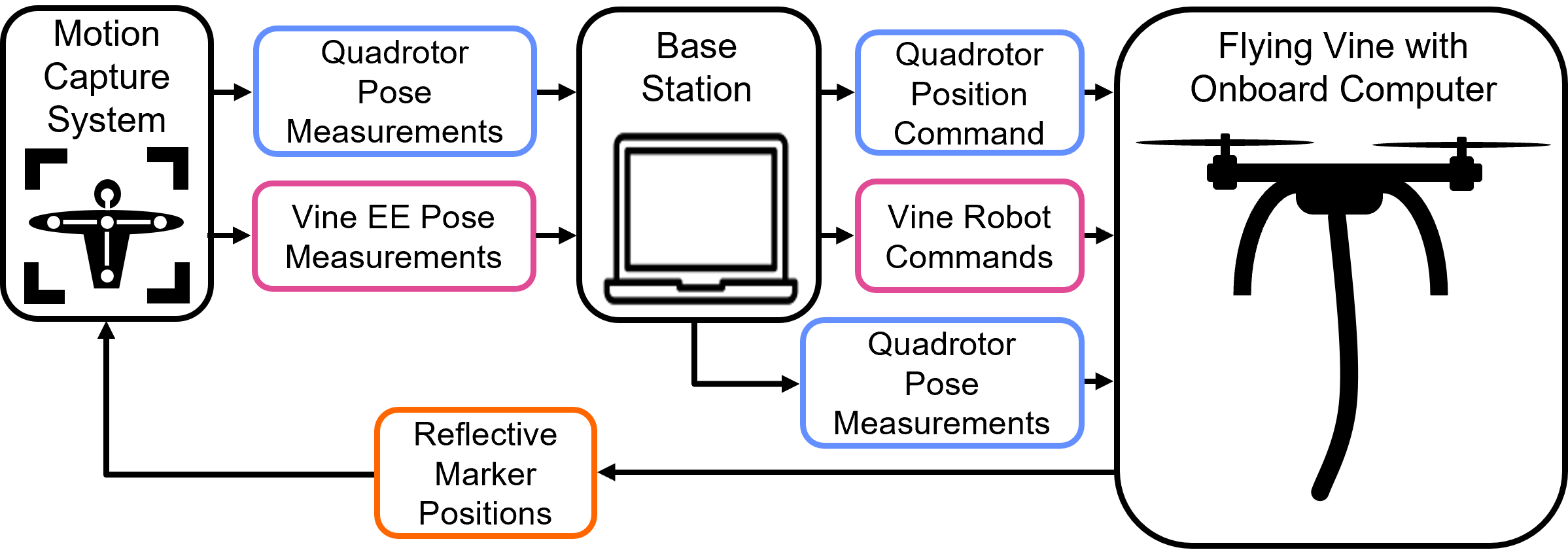}
    \vspace{-7mm}
    \caption{High-level illustration of data flow for operating the flying vine.}
    \label{fig:comms}
    \vspace{-4.5mm}
\end{figure}

We adapt an existing sensing and communication infrastructure (\href{https://github.com/StanfordMSL/TrajBridge}{\textit{TrajBridge}}) for the flying vine (Fig.~\ref{fig:comms}). The motion capture system runs OptiTrack  software (Motive 2.0) to take in camera data, extract reflective marker positions, and publish pose measurements for the quadrotor and end effector. Motive streams these measurements on a wireless network via the Virtual Reality Peripheral Network (VRPN) protocol. 

The base station (Lenovo Legion laptop) ports motion capture messages onto a ROS network with the \textit{vrpn\_client\_ros} package and also stores the pose measurements for offline use (model fitting and performance analysis). The base station also publishes quadrotor position commands at 20~Hz. The sequence of commands sent by the base station is optimized offline and published open-loop at runtime, although the quadrotor's flight controller runs closed-loop control on the position commands it receives. Finally, the base station sends vine robot commands (growth and pressure commands).

The onboard quadrotor computer (UP Board) subscribes to ROS messages from the base station for the current quadrotor pose as well as the current quadrotor position commands, and then passes these to the flight controller for closed-loop control on quadrotor position. It also subscribes to vine robot commands and forwards them to the vine robot’s microcontroller. 

% ============================
% ============================

% -------------------------------------------------------------

\section{Modeling and Control}

A common control task for robot arms is positioning the end effector. For the flying vine, we move the end effector by sending position commands to the quadrotor, and the motion of the quadrotor causes motion of the vine robot and thus its end effector. Changing the length and pressure of the vine robot changes its kinematics as well as dynamic parameters such as its effective stiffness and damping, but these changes happen on much longer timescales compared to quadrotor motion. Thus, we first discuss modeling assuming a fixed vine pressure and length (Sec.~\ref{sec:model}) and then discuss a model that allows for time-varying pressure and length (Sec.~\ref{sec:bilinear}).

We considered several options for modeling flying vine behavior. Traditional aerial robot arms can directly use rigid body dynamics, and these techniques can be applied to a continuum arm by abstracting the continuum arm as a finite number of links\cite{Szasz2022}. A further simplifying assumption would be to abstract the flying vine as a quadrotor with a cable-suspended load and draw from existing slung-load literature\cite{Foehn2017, autotrans}. Data-driven models are an alternative to these physics-based techniques\cite{Centurelli2022, inertial_dynamics}. We hypothesize that any of these methods would work well, although we anticipate challenges with rigid body approximations in modeling a time-varying vine length, and we expect lower model accuracy with slung-load approximations as the vine body pressure increases. Thus, for this initial work, we chose to build on\cite{inertial_dynamics} because that data-driven model: (1) is simple yet effective, with a straightforward parameter fitting procedure, (2) was used previously to successfully control a physical inflated-beam robot arm, and (3) is easily applied to the flying vine in a way that also simplifies the control strategy (Sec.~\ref{sec:trajopt}).

\subsection{Data-Driven Model}\label{sec:model}

\begin{figure}
\vspace{2.5mm}
    \centering
    \input{FinalFigures/pretzel.tikz}
    \vspace{-5mm}
    \caption{Example flying vine motion. We overlay the quadrotor position command (dotted), actual quadrotor position (dashed), and end effector position (solid) to show that there are offsets between these three signals that need to be modeled.}
    \label{fig:tracking_error}
    \vspace{-4.5mm}
\end{figure}
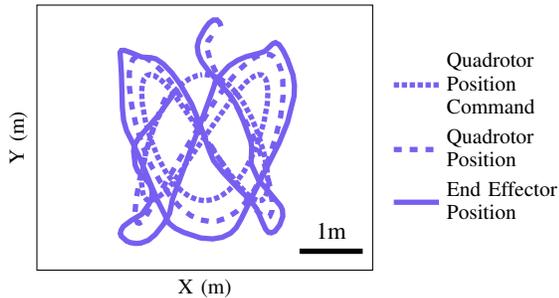

In our data-driven approach to modeling, we treat fitting a dynamics model as a regression problem in which we aim to predict the subsequent state given current and past information about the state and control input. We have access to quadrotor position and orientation as well as end effector position and orientation through motion capture measurements, although we exclude end effector orientation given its poor signal-to-noise ratio. Thus, we define the state of the flying vine as a concatenation of the quadrotor position ($\mathbb{R}^3$), quadrotor orientation (quaternion), and end effector position ($\mathbb{R}^3$). We observed that the scalar component of the quaternion was always near 1 and that excluding it from the state vector did not hinder model accuracy. This is likely because the flying vine quadrotor does not exhibit extreme rotation angles, and the vector component of a quaternion is proportional to the axis-angle representation under small-angle assumptions. This exclusion yields the state vector $\*x \in \mathbb R^9$. Our control input $\*u \in \mathbb R^3$ is the $xyz$ quadrotor position command. The control input $\*u$ is a desired quadrotor position, which is different from the actual quadrotor position given in $\*x$ because the quadrotor's onboard controller exhibits some tracking error (Fig.~\ref{fig:tracking_error}). 
% ============================
% ============================
We define the general discrete dynamics function

\begin{equation}
\begin{aligned}
\*x^{(k+1)} &\coloneq f(\*x^{(k)}, \*x^{(k-1)}, \*x^{(k-2)}, \*u^{(k)}) \\
&\coloneq f(\*z^{(k)}, \*u^{(k)}), 
\end{aligned}\label{eq:dynamics}
\end{equation}

\noindent where $(k)$ is the current timestep, and $\*z$ is referred to as the augmented state. By including the two previous states in the function's inputs, we provide the fit model indirect access to system velocity and acceleration. A simple yet effective model choice is to assume a linear function, $f(\*z, 
\*u) \coloneq A\*z + B\*u$, which worked well in~\cite{inertial_dynamics} to model a ceiling-mounted soft robot arm. The authors also provide details on how to fit the elements of A and B using experimental data and least squares. 

This model captures general flying vine motions, but the model accuracy for end effector height suffers in swinging motions where the end effector height deviates significantly from its nominal value. To address this, we introduce higher-order terms to our model for predicting the end effector height. In particular, we introduce squared and bilinear terms from the concatenation $(\*z, \*u)$, as well as an offset term. As an example, the expression for a 2-element augmented state and a scalar control input would be
\begin{equation}
\begin{aligned}
f(\*z, u) \coloneq a_0 &+ a_1 z_1 + a_2 z_2 + a_3 u \\
&+ a_4 z_1^2 + a_5 z_2^2 + a_6 u^2 \\
&+ a_7 z_1 z_2 + a_8 z_1 u + a_9 z_2 u,
\end{aligned}
\label{eq:tip_height}
\end{equation}
\noindent where $\*a$ contains the model parameters. This function is nonlinear with respect to $\*z$ and $\*u$, but it is linear with respect to $\*a$, so we can still use least squares to fit the values of $\*a$. We use MATLAB’s \textit{fitlm} function to perform this regression. Note that this modification is only applied to the model for end effector height. The remaining states are modeled with $A\*z + B\*u$.

The training data used to fit the model comes from experimental flight data that is representative of the motions we expect in the control tasks. For the flying vine, we would like the end effector to follow both low and high speed trajectories, so we handcrafted a few command trajectories that varied in speed and had position commands that were distributed across the primary working space of the flight room. Fig.~\ref{fig:tracking_error} is an example from the training data. 

\subsection{Bilinear Interpolation of Time-Varying Parameters}
\label{sec:bilinear}

The process outlined above yields a dynamic model that captures the behavior of the flying vine at a single pressure and length configuration. The flying vine has qualitatively similar behavior if the pressure or length is changed, but the model accuracy would suffer. To address this, we use bilinear interpolation\cite{Kirkland2010} with 4 models of the vine: empty short~(ES), inflated short~(IS), empty long~(EL), and inflated long~(IL). We collect 4 separate training datasets for each configuration and fit one model per dataset. With these 4 models, we can interpolate for any vine configuration that falls within the pressure and length bounds of the 4 models, which is 0-0.4~kPag and 0.7-1~m respectively.

\subsection{Offline Trajectory Optimization}\label{sec:trajopt}

We achieve end effector positioning with offline trajectory optimization, and we send the optimized commands open-loop at runtime. The quadrotor still performs real-time feedback control on the received position command. The general form we use for trajectory optimization is:
\begin{mini}
    {Z,U}
    {\sum_{k=1}^{N} 
    \norm{\*z^{(k)} - \bar{\*z}^{(k)}}^2_Q + 
    \sum_{k=1}^{N-1} 
    \norm{\*u^{(k)} - \bar{\*u}^{(k)}}^2_R}
    {}{}
    \addConstraint{\*z_{1:9}^{(k+1)}}{= f(\*z^{(k)}, \*u^{(k)}), \quad}{k=1,\dotsc,N-1}
    \addConstraint{\*z_{10:27}^{(k+1)}}{= \*z_{1:18}^{(k)}, \quad}{k=1,\dotsc,N-1}
    \addConstraint{\*u_{\text{min}}}{\leq \*u^{(k)} \leq \*u_{\text{max}}, \quad}{k=1,\dotsc,N-1}
    \addConstraint{\*z^{(1)}}{= \*z_{\text{rest}}, \quad}
    \label{opt:trajopt}
\end{mini}
\noindent where $Z \coloneq \*z^{(1:N)}$ and $U \coloneq \*u^{(1:N-1)}$ are the optimized state and control trajectories, $\bar{\*z}^{(1:N)}$ and $\bar{\*u}^{(1:N-1)}$ are the reference state and control trajectories, and $Q$ and $R$ are weight matrices on state and control deviation, respectively. The first two constraints define the discrete dynamics. The augmented state $\*z \in \mathbb R^{27}$ contains the current and two previous states, so the first 9 elements of $\*z^{(k+1)}$ come from Eq.~\ref{eq:dynamics}, and the last 18 elements, $\*z_{10:27}^{(k+1)}$, are the first 18 elements of $\*z^{(k)}$. The third constraint is for control limits, and the last constraint enforces the trajectory to start from a nominal rest configuration (quadrotor and end effector at x=0, y=0).

A simple choice for $\bar{\*z}^{(1:N)}$ is to set the quadrotor positions to have a fixed offset from the desired end effector positions (i.e. assume the quadrotor and end effector are rigidly attached), although we make some refinements for some of our experiments (Sec.~\ref{sec:results}). Defining a reference control input, $\bar{\*u}^{(1:N-1)}$, is also straightforward because the control input of our dynamic model is a quadrotor position command rather than something lower level such as thrust and torque or even motor commands. As an example, if the task was to move the end effector in a straight line, the reference control input could also be a straight line with a z-offset corresponding to the z-offset of the quadrotor with respect to the end effector. 

We use an optimizer written in Julia \href{https://github.com/thowell/IterativeLQR.jl}{\textit{(IterativeLQR.jl)}}, which uses an iterative linear quadratic regulator (iLQR) for handling nonlinear dynamics and an augmented Lagrangian framework for handling constraints \cite{howell2019altro}. 
 
% -------------------------------------------------------------
\section{Results}\label{sec:results}

We consider two types of control tasks to quantify the performance: (1) tracking a lemniscate path ($\infty$) with the end effector and (2) swinging the end effector to a specific position target. We quantify performance for each of the 4 vine configurations: empty short~(ES), inflated short~(IS), empty long~(EL), and inflated long~(IL). We then demonstrate tracking of a lemniscate path while the vine grows and inflates, i.e. a transition from ES to IL. This requires the bilinear interpolation described in Sec.~\ref{sec:bilinear}.

\subsection{End Effector Tracking of a Lemniscate Path}\label{sec:fig8}

% ============================
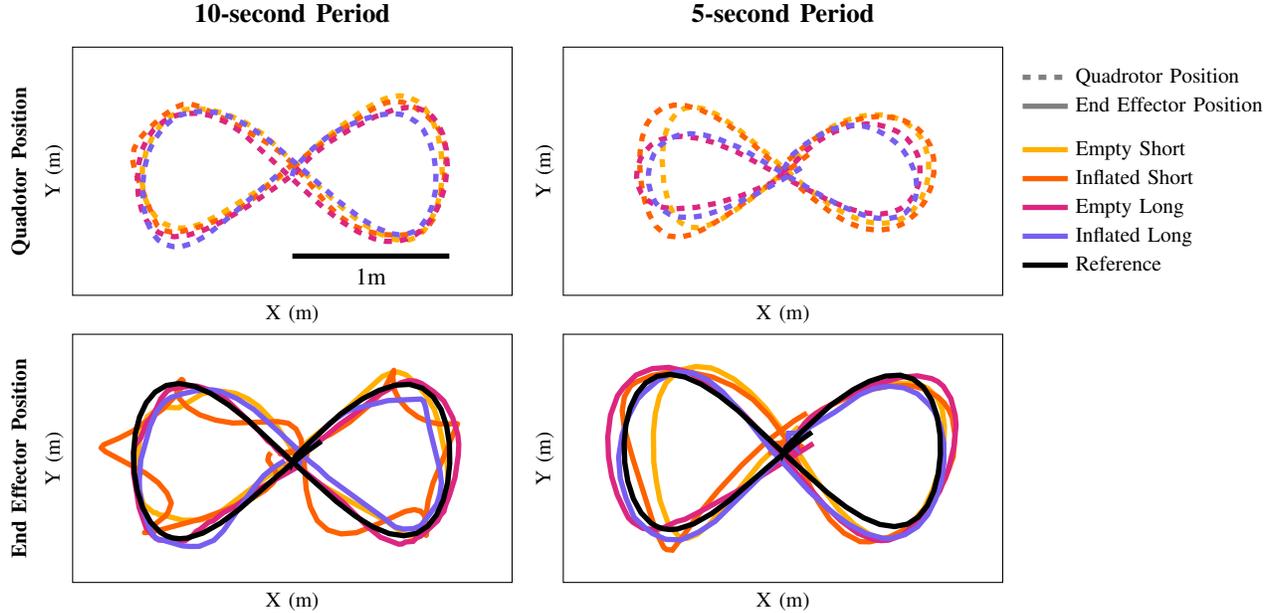
\begin{figure*}[tp]
\vspace{2.5mm}
\centering
\input{FinalFigures/2by2.tikz}
\vspace{-4mm}
\caption{Comparison of results for lemniscate path end effector tracking with the 4 vine configurations (empty short, inflated short, empty long, and inflated long). The top plots show quadrotor motion, and the bottom plots show end effector motion. The left plots show the low-speed motion, and the right plots show the high-speed motion. The mean distance (3D Euclidean) between the actual and reference end effector position trajectory ranges from 5-9 cm, which is small compared to the width of the lemniscate path (2~m).}
\label{fig:2by2fig8}
\vspace{-4.5mm}
\end{figure*}
% ============================

In this first task, we define a lemniscate end effector reference trajectory:
\begin{equation}
\begin{aligned}
    x &= \sin \frac{2 \pi t}{T}  \\
    y &= \cos \frac{2 \pi t}{T} * \sin \frac{2 \pi t}{T} \\
    z &= 1.5 - l_\text{vine}
\end{aligned}
\end{equation}
where $l_\text{vine}$ is the length of the vine and $T$ is the time to complete a single lap. We measure flying vine performance at two periods, $T=5$ and $T=10$, and compare the difference in behavior. We also modify the reference trajectory by adding an initial ramp-up time of 2 and 3 seconds for the slow and fast trajectories, respectively, to reduce the initial acceleration required to begin motion.

We require reference trajectories for the rest of the state as well as the control input in order to use our trajectory optimization framework (Eq.~\ref{opt:trajopt}). For simplicity, we set the quadrotor position reference to match the end effector reference except with a z-offset equivalent to the vine length, and we set the quadrotor reference orientation to have a rotation angle of zero. 

Defining a reference control trajectory is straightforward since our model's control input is a position command for the quadrotor. The simplest option would be to have the reference control match the reference state for the quadrotor's position, $\bar{\*u}^{(1:N-1)} \coloneq \bar{\*z}_{\text{QR,xyz}}^{(1:N-1)}$. However, we know a priori that there is some lag and offset from the quadrotor position command to the end effector position (Fig.~\ref{fig:tracking_error}). Thus, we introduce scaling and time-shift parameters to improve the control reference:
% \vspace{-2mm}
\begin{equation}
\begin{aligned}
    \bar u_x^{(k)} &= \alpha_x * \bar z_\text{QR,x}^{(k+\alpha_t)} \\
    \bar u_y^{(k)} &= \alpha_y * \bar z_\text{QR,y}^{(k+\alpha_t)}. %\\
    % \bar u_z &= \alpha_z \bar z_\text{quadrotor z}
\end{aligned}
\end{equation}
We hand-tune the parameters with trial-and-error in simulation. That is, we choose a set of values for $\bm{\alpha}$, generate $\bar{\*u}^{(1:N-1)}$, simulate the flying vine motion for $N$ timesteps, and plot the end effector motion against the reference. The purpose of this step is to quickly find a reasonable control reference for the trajectory optimizer to improve on, so we selected the best set of parameters after about five attempts. We set $\alpha_t = 10$, $\alpha_x = 0.9$, and $\alpha_y = 1$ or $\alpha_y = 0.6$ for the slow or fast speeds, respectively.

We use diagonal weight matrices $Q$ and $R$. We prioritize end effector tracking by setting a weight of 20 for the indices of $Q$ corresponding to the end effector position and setting the others to 1. The control limits are set to $\pm 3$~meters, although the optimized trajectory stayed within the limits. 

Fig.~\ref{fig:2by2fig8} shows the results for the two different speeds and compares the performance for each of the four vine configurations (ES, IS, EL, and IL). Comparing the slow and fast trajectories, we see the quadrotor traces a smaller pattern when tracking at a higher speed. This makes sense given that we expect higher centrifugal forces on the end effector at higher speeds, and this trend is more pronounced for the longer vine configurations (EL and IL) given their kinematics. The mean distance (3D Euclidean) between the actual and reference end effector position trajectory ranges from 5-9~cm, which is small compared to the width of the lemniscate path (2~m). The exception is the IS configuration with a mean distance of 12~cm, which exhibits noticeable oscillations for the low-speed trajectory. The IS vine has the highest natural frequency of the configurations and likely the strongest dynamic coupling between quadrotor and end effector. We hypothesize that at the higher speed, vine behavior is primarily dictated by the quadrotor's large-scale movements, but at a lower speed, the IS vine can more easily build up energy in its swinging mode.

\subsection{Swinging to a Target End Effector Position}

In the previous lemniscate tracking task, achieving a desired end effector trajectory can be achieved with little bending of the vine robot, so here we define a swinging task that requires more vine bending. This demonstrates that our model and control framework can handle more complex behaviors, and highlights the added workspace that comes with using a flexible robot arm. Moreover, the bending is achieved by leveraging the high maneuverability of the quadrotor rather than having to use additional control inputs on the robot arm, which is the case on a traditional aerial robot arm. We envision this swinging capability to be useful when we need to reach a target end effector position under significant position constraints. For example, a swing maneuver could be used for initial entry of a vine robot in a pipe inspection task, or for hooking a sensor on a tree branch while keeping the quadrotor at a safe distance. 

In this work, we simplify the task to ``kicking" a beach ball to serve as an initial example for using our modeling and control framework for swinging behaviors~(Fig.~\ref{fig:summary}). In order to yield a swinging trajectory, we enforce narrow control limits on the y-command and z-command of the quadrotor. Thus, the optimized control relies on its x-command to coordinate lateral motion of the quadrotor to achieve swinging of the vine. 

For this task, we define a target end effector position in the $xz$\nobreakdash-plane $(x_\text{target}, z_\text{target})$. The end effector should have some momentum when it reaches the target position, so we define a target end effector position for the prior state, which indirectly sets an end effector target velocity. The target positions are selected such that the end effector velocity is forwards and upwards in the $xz$ directions. To achieve this, we augment the trajectory optimization in Eq.~\ref{opt:trajopt} with the following equality constraints: 
\begin{equation}
\begin{aligned}\label{eq:equality_constraints}
(\*z^{(N/2)}_{QR,x}, \*z^{(N/2)}_{QR,z}) &= (x_\text{target}, z_\text{target})\\
(\*z^{(N/2-1)}_{QR,x}, \*z^{(N/2-1)}_{QR,z}) &= (x_\text{target}-0.1, z_\text{target}-0.1).
\end{aligned}
\end{equation}
\noindent This constraint is applied to the halfway point of the trajectory ($N/2$) so that the optimized trajectory includes the ``swing-up" as well as the return to the starting position. Finally, we add x- and z-limits to the quadrotor position to simulate a scenario when the quadrotor needs to maintain some distance from the target position or obstacles in the environment.

We define weight matrices $Q \coloneq I$ and $R \coloneq 10I$. Tracking the state reference is not the focus of this control task, so we use a lower scaling for $Q$ and let the equality constraints (Eq.~\ref{eq:equality_constraints}) drive the optimization. We set the control limits to be $(\pm 2, \pm 0.1, 1.5 \pm 0.1)$.

The state and control reference trajectories come from an early exploratory experiment on generating swinging motions with hand-made control trajectories. From physical intuition, the quadrotor should move forwards and backwards in the x-direction to achieve a swinging motion, so we explored a control trajectory with a trapezoidal shape (for $x$ vs. $t$) because of its simplicity and lack of sudden jumps in position command. However, we believe other options would also have worked well. The experiment we chose to use for the state and control reference achieved qualitative swinging but did not pass through the target position. Furthermore, we use the same references for all 4 optimizations (one per vine configuration), which verifies that the reference trajectories do not need to be close to optimal. It was convenient to use existing data for the reference trajectories, but a similar procedure based in simulation (as in Sec.~\ref{sec:fig8}) should also work well.

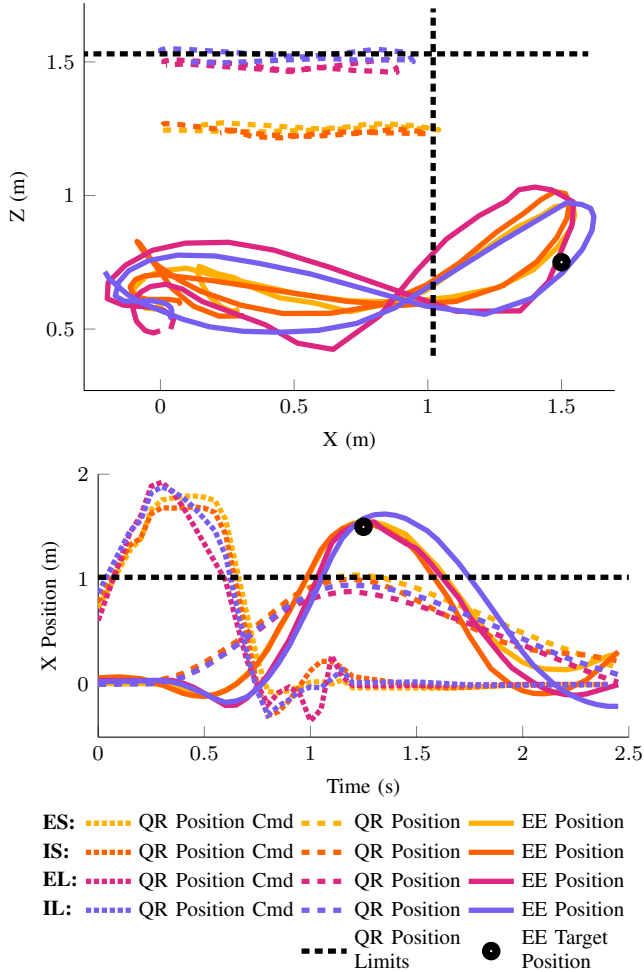
\begin{figure}
\vspace{2.5mm}
    \centering
    \input{FinalFigures/kick_xz.tikz}
    \input{FinalFigures/kick_ux.tikz}
    \vspace{-7mm}
    \caption{Results on hardware for swinging to a target end effector (EE) position, shown with a black circle. We add x- and z-limits to the quadrotor (QR) position to simulate a scenario when the quadrotor needs to maintain some distance from the target position or obstacles in the environment. Limits are shown with black dashed lines. Results from all 4 vine configurations are overlaid: empty short~(ES), inflated short~(IS), empty long~(EL), and inflated long~(IL). (Top) Comparison of motion in the $xz$\nobreakdash-plane. The plots for the short configurations were shifted down to align the target positions. 
    (Bottom) Comparison of quadrotor x-position command, actual quadrotor x-position, and end effector x-position over time. The target should be reached at 1.25 s.}
    \label{fig:kick}
    \vspace{-4.5mm}
\end{figure}

We deployed the optimized control trajectories on the physical flying vine in free space to measure performance and then repeated one experiment with a beach ball as a demonstration (Fig.~\ref{fig:summary}). Fig.~\ref{fig:kick} shows the quadrotor and end effector motions in the $xz$\nobreakdash-plane, and the target end effector position is shown with a black circle. The mean distance (3D Euclidean) between the actual and modeled end effector position trajectory ranges from 13-16~cm, which is low compared to the range of end effector motion (1.5~m in the x-direction). While this tracking error is larger than that of the lemniscate path, this is reasonable given the increased difficulty of this task. 
% ============================
% ============================
Fig.~\ref{fig:kick} also shows the time series of quadrotor x-position command, quadrotor x-position, and end effector x-position. Plotting these signals over time highlights the time delays between the peaks of the three signals, which further emphasizes the need to model the tracking error of the quadrotor's flight controller as well as the relative motion of the vine with respect to the quadrotor. We note that there are a couple instances of exceeding quadrotor position limits, although the deviations are small (2~cm). We hypothesize this is due to minor modeling error combined with the decision to send position commands open-loop, and we are interested in closed-loop control for future work.
% ============================
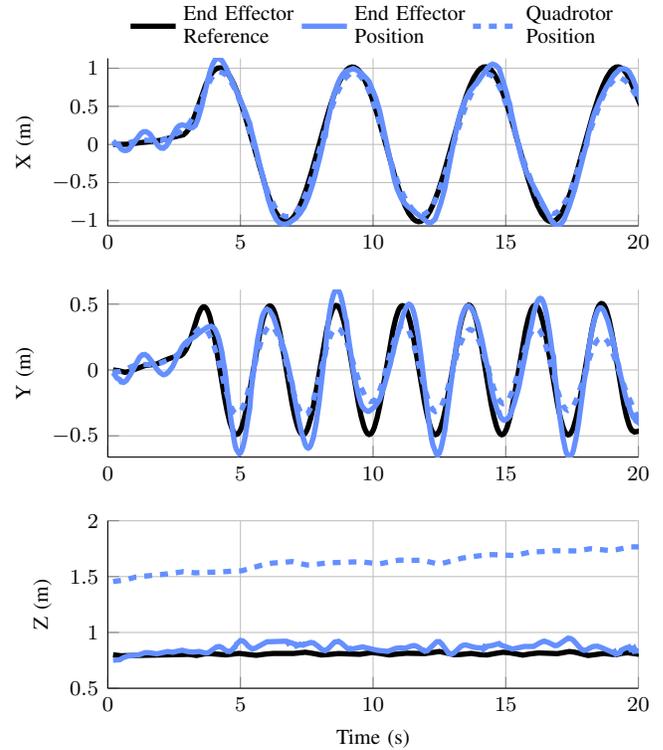
\begin{figure}
\vspace{2.5mm}
    \centering
    \input{FinalFigures/growing_xyz.tikz}
    \vspace{-7mm}
    \caption{Time series results from tracking a lemniscate path ($\infty$) as the vine robot grows and inflates. The quadrotor position must increase over time to accommodate the increasing vine length and maintain a fixed end effector height.}
    \label{fig:growing}
    \vspace{-4.5mm}
\end{figure}
% ============================
\subsection{Demonstration with Time-Varying Parameters}

We provide a demonstration with the bilinear interpolated model proposed in Sec.~\ref{sec:bilinear} by optimizing a 20-second trajectory in which the vine grows 30~cm and goes from empty to inflated (i.e. a transition from ES to IL). This is akin to deploying the vine robot from a stowed configuration. The end effector reference trajectory is a lemniscate path with a constant height, so the quadrotor must increase its height over time to accommodate the increasing vine length. The trajectory optimization setup is the same as in Sec.~\ref{sec:fig8}, except the state and control references have linearly increasing values for quadrotor height, and the dynamics constraint uses the interpolated dynamics function. We run the optimized trajectory on the physical flying vine and show the time series positions in Fig.~\ref{fig:growing}. The mean distance (3D Euclidean) between the actual and modeled end effector position trajectory is 15~cm, which is low compared to the range of end effector motion (2~m in the x-direction). These results underscore the value of simple yet effective modeling techniques such as bilinear interpolation in addressing the challenges of controlling soft robot arms with time-varying dynamics.

% -------------------------------------------------------------
\section{Conclusion}

In this work, we present the design of a soft aerial robotic arm that combines a compact, lightweight, soft growing arm with a small quadrotor. The flying vine design leverages vine robot compliance and capacity for length change to achieve a small stowed configuration and a large end effector workspace. While the vine robot is underactuated, the flying vine leverages quadrotor maneuverability to control the motion of its end effector. Experimental results show that the flying vine with the proposed modeling and control framework can achieve compelling behaviors, and lays the necessary groundwork for developing flying vine technology. For future hardware iterations, we are interested in outdoor flight capability and end effector designs for different manipulation tasks. In terms of the control framework, we are most interested in implementing real-time feedback control on end effector position, which opens doors for disturbance rejection as well as operation with an unknown or time-varying payload mass. 

\section*{Acknowledgements}
% Hidden for double-blind review
The authors thank Keiko Nagami, Jun En Low, and Kenneth Hoffman for quadrotor advice, Prof. Mac Schwager and his Multi-Robot Systems Lab for lending a quadrotor, and Prof. Zachary Manchester for insightful discussions.
%%%%%%%%%%%%%%%%%%%%%%%%%%%%%%%%%%%%%%%%%%%%%%%%%%%%%%%%%%%%%%%%%%%%%%%%%%%%%%%%
\bibliographystyle{IEEEtran}
% \IEEEtriggeratref{13}
\bibliography{paper}

\end{document}

%% file: FinalFigures/pretzel.tikz
\definecolor{mycolor1}{rgb}{0.47059,0.36863,0.94118}%
\begin{tikzpicture}[font=\footnotesize]
\pgfplotsset{ticks=none}

\begin{axis}[%
axis equal,width=.7\columnwidth,
height=.75*0.7887\columnwidth,
at={(0.758in,0.481in)},
% scale only axis,
% xmin=-2.20618934870015,
% xmax=2.28390533940822,
xmin=-2.1,
xmax=2.1,
xlabel={X (m)},
ymin=-2.1,
ymax=2.1,
ylabel={Y (m)},
axis background/.style={fill=white},
legend style={at={(1.03,0.5)}, anchor =west, legend cell align=left, align=left, draw=none}
]
\draw[line width=2pt, black] (axis cs:1.5,-1.8) -- (axis cs:2.5,-1.8); % Thick black line
    \node[above] at (axis cs:2,-1.75) {\small 1m}; % Label below the line

\addplot [color=mycolor1, densely dotted, line width=2.0pt]
  table[row sep=crcr]{%
0	1\\
0.125	0.982\\
0.242224118738427	0.933470573329099\\
0.368	0.844\\
0.482	0.729\\
0.615902047203042	0.544033137740661\\
0.743	0.309\\
0.844	0.0629999999999999\\
0.937	-0.249\\
0.989	-0.536\\
1	-0.729\\
0.982	-0.876\\
0.951	-0.951\\
0.905	-0.992\\
0.844	-0.998\\
0.771	-0.969\\
0.653	-0.876\\
0.484082283996963	-0.687545013774065\\
0.208	-0.309\\
-0.518	0.729\\
-0.682273049241473	0.902528700875085\\
-0.797	0.982\\
-0.866	1\\
-0.922	0.982\\
-0.962802822276912	0.930345061015405\\
-0.98893499827527	0.844297150741622\\
-1	0.729\\
-0.989	0.536\\
-0.937	0.249\\
-0.866	0\\
-0.743	-0.309\\
-0.621	-0.536\\
-0.482	-0.729\\
-0.368	-0.844\\
-0.249	-0.93\\
-0.125	-0.982\\
0	-1\\
0.125	-0.982\\
0.249	-0.93\\
0.368	-0.844\\
0.477132494709617	-0.73452527627557\\
0.621	-0.536\\
0.743	-0.309\\
0.844	-0.0629999999999999\\
0.937	0.249\\
0.989	0.536\\
1	0.729\\
0.982	0.876\\
0.951	0.951\\
0.905	0.992\\
0.844	0.998\\
0.771	0.969\\
0.653	0.876\\
0.482	0.685\\
0.208	0.309\\
-0.518	-0.729\\
-0.653	-0.876\\
-0.771	-0.969\\
-0.866	-1\\
-0.922	-0.982\\
-0.963	-0.93\\
-0.982327186111504	-0.874504292061695\\
-1	-0.729\\
-0.989	-0.536\\
-0.937	-0.249\\
-0.844	0.0629999999999999\\
-0.743	0.309\\
-0.621	0.536\\
-0.482	0.729\\
-0.368	0.844\\
-0.249	0.93\\
-0.125	0.982\\
-0.042	0.998\\
-0.042	0.998\\
};
\addlegendentry{Quadrotor \\Position \\Command}

\addplot [color=mycolor1, dashed, line width=2.0pt]
  table[row sep=crcr]{%
0.123294639462473	1.5621631109908\\
0.0914668549968807	1.58074068826637\\
0.0532808246479244	1.53855184745018\\
0.0369055723785772	1.47675332370452\\
0.0408298312216737	1.39022590189336\\
0.0804176945164243	1.24388653966467\\
0.158894767933918	1.07026529185657\\
0.315756476913451	0.807528265303975\\
0.627829357067415	0.293409104127432\\
0.806619145453759	-0.0665541022267317\\
0.944707302978411	-0.427647844152224\\
1.0255620469797	-0.746487739739446\\
1.0671053682345	-1.01361912868533\\
1.063030921423	-1.16231751308003\\
1.02790432384792	-1.2485990615251\\
0.982254682255877	-1.26555777064636\\
0.921857432344223	-1.24482014995003\\
0.842298483530206	-1.17851695173845\\
0.678150877393459	-0.984736379438933\\
0.422402171065449	-0.634850054880252\\
0.0833005614995004	-0.103166669865002\\
-0.499710385395609	0.829329460765808\\
-0.674736966852038	1.04702262095132\\
-0.830771106902423	1.19736157747418\\
-0.938584258791593	1.26648192178752\\
-1.02868819037552	1.29197614164374\\
-1.08073085176703	1.2858959629358\\
-1.13770801090254	1.24170849402298\\
-1.17642164489065	1.15139502001549\\
-1.19517541720378	1.02250719866309\\
-1.19239788800874	0.791739605789111\\
-1.15713342478122	0.511823537793958\\
-1.09033780754504	0.20721074316107\\
-0.99290984830784	-0.110560010031964\\
-0.857929456542072	-0.428173395724879\\
-0.697049895920628	-0.713641873924968\\
-0.516524492668704	-0.966073033244835\\
-0.373483606795703	-1.11995229465525\\
-0.220452849172855	-1.23520890824877\\
-0.0672273378074379	-1.30386215229605\\
0.0942835973839629	-1.32955880708838\\
0.199442461279524	-1.32107524809597\\
0.355796450442367	-1.26762776406894\\
0.501970337322572	-1.17118862554715\\
0.641068259623893	-1.03170370273546\\
0.771141933134561	-0.851595372070697\\
0.887083899186205	-0.638819392037246\\
1.01395282145516	-0.329799865284181\\
1.12110668249912	0.00224927700656119\\
1.20154086489997	0.339224343868641\\
1.2390231060917	0.647793757041261\\
1.23490354869359	0.909999806317483\\
1.20551166848858	1.07460203982808\\
1.15509547286477	1.20302268621693\\
1.08665758695461	1.28476896565409\\
1.02956909609822	1.31609383283594\\
0.933118371554616	1.32460711589718\\
0.816206144103033	1.28568119511449\\
0.689460506878393	1.20474452347728\\
0.550886014424268	1.0816182670216\\
0.361102563645402	0.861337055643453\\
0.160465536445219	0.580084785374856\\
-0.185368045297812	0.0187826999326557\\
-0.978781446464589	-1.32506765742943\\
-1.03585417873196	-1.35612545170973\\
-1.06493271839314	-1.34920594701432\\
-1.0938788911021	-1.29861048585372\\
-1.10493070877029	-1.19678069997958\\
-1.09039276249653	-1.00160301556066\\
-1.03312096728894	-0.671849555770382\\
-0.916668520374919	-0.186982190744412\\
-0.810884621528296	0.162685138642934\\
-0.6806457216295	0.502217491476566\\
-0.569191634311799	0.73917066123417\\
};
\addlegendentry{Quadrotor\\ Position}

\addplot [color=mycolor1, line width=2.0pt]
  table[row sep=crcr]{%
0.257551101110614	1.76320797751561\\
0.237427980241202	1.8203023144671\\
0.192742318827477	1.86543428426827\\
0.133488521975333	1.86318391551734\\
0.0665618215552584	1.8138318533358\\
-0.00281811501638818	1.72090362437191\\
-0.0900401123491454	1.56077448159672\\
-0.118976329306113	1.44840113583766\\
-0.103425472810216	1.32670386169285\\
-0.0514885090859725	1.20850365405607\\
0.0613773957698467	1.05541453362351\\
0.405480561505442	0.665035177091617\\
0.491956731982426	0.528751207337463\\
0.617857425405465	0.25072991201719\\
0.729094160432219	-0.0585184807173627\\
0.743120162944537	-0.153883198610997\\
0.732991967658767	-0.411346390265424\\
0.850478495951539	-0.67199162718281\\
1.02562704543544	-0.870886050835883\\
1.19962145458424	-1.08223232457998\\
1.22874994071514	-1.14321433200709\\
1.23556216222926	-1.24690771766886\\
1.18622207331752	-1.38374126419625\\
1.11081563498346	-1.4844451136603\\
1.05272116514454	-1.50848895280923\\
1.00631543404963	-1.50070248669401\\
0.963789163021283	-1.45105714557697\\
0.903306617117368	-1.31868642954235\\
0.824331764485511	-1.13324098204049\\
0.73121194886361	-1.0040276720166\\
0.532343364690731	-0.809442784565505\\
0.288068863262837	-0.551379456556624\\
0.140259630163809	-0.36787388840489\\
0.0730067649774429	-0.208602211480855\\
-0.0732040928525255	0.202281263260326\\
-0.209239348526562	0.535646277284768\\
-0.390159013878182	0.850272457847505\\
-0.523484773011882	1.03023683149308\\
-0.655901765528326	1.1665935683216\\
-0.769591779743846	1.24561584543278\\
-0.849050020287184	1.26687242225427\\
-0.994362412348331	1.27607841503099\\
-1.16164661059099	1.34295744941077\\
-1.27390247936794	1.38366261122456\\
-1.32156779444267	1.38123098082851\\
-1.34231486495858	1.34935501633621\\
-1.33270742060312	1.22572272832706\\
-1.31969493705512	1.00086169914935\\
-1.3044990130514	0.473799244648432\\
-1.28442833002176	0.271066876083059\\
-1.23742677213086	0.108214561993503\\
-1.13232295082282	-0.100888578112355\\
-1.03041221570222	-0.326402227371192\\
-0.742726372228947	-1.00625405303381\\
-0.567465251690197	-1.35531198799244\\
-0.466221211088099	-1.43675574859459\\
-0.332718253404021	-1.50226593799674\\
-0.0647151888727011	-1.58968627974336\\
0.125182395743764	-1.60496888465627\\
0.278875564657117	-1.57808411600145\\
0.41035145415726	-1.51384371950111\\
0.497520260631335	-1.42671174308028\\
0.575525462765288	-1.29058782460646\\
0.738729057262362	-0.958785460828051\\
1.08996777324188	-0.255764665770948\\
1.17476658552432	-0.0904818159111533\\
1.3094876618669	0.242425128619744\\
1.41415771238773	0.654751239807811\\
1.42003085566664	0.827105204664902\\
1.39551003378127	0.996818712926779\\
1.30938876375026	1.23690893996126\\
1.24246205321918	1.37671462794471\\
1.17595166726546	1.47338037334633\\
1.11879629756131	1.50605654355915\\
1.08035314296207	1.49110455913886\\
0.948186441254069	1.36605779369554\\
0.857438950373557	1.33619432576988\\
0.653958567222284	1.28265489774893\\
0.521189047292395	1.22655195220704\\
0.378846263136499	1.13330596037443\\
0.173233986805264	0.910476336291975\\
0.140289203932216	0.797510540126858\\
0.0675373298428781	0.592202702062692\\
-0.511679351655058	-0.975268982035761\\
-0.556286906692831	-1.19568895825132\\
-0.651940296296672	-1.36910797795516\\
-0.785537318501067	-1.53689763845239\\
-0.89544812515465	-1.62788414493635\\
-1.0159666175402	-1.67188411858525\\
-1.0916376186523	-1.67594684877204\\
-1.17382585061197	-1.664451645754\\
-1.23691066869781	-1.63879532038816\\
-1.32224735180331	-1.56223461447153\\
-1.33828105830305	-1.51991678060049\\
-1.32530884935114	-1.46693387938436\\
-1.28970411963056	-1.40670811520697\\
-1.15908443780428	-1.30819105335292\\
-0.997346347201704	-1.21399955838081\\
-0.886995963629652	-1.05729555574395\\
-0.845696011281635	-0.969124399390123\\
-0.833587052971745	-0.882615262078474\\
-0.845893521525581	-0.801297365031553\\
-0.875610538090947	-0.729473657021396\\
-0.993922525043423	-0.602272137273047\\
-1.06333540291757	-0.530109580270862\\
-1.10213045258932	-0.436974300108754\\
-1.12710163656994	-0.210117001629406\\
-1.106581416045	-0.104358824597685\\
-1.04267450017646	-0.000248937090229884\\
-0.852302329253089	0.219668213320346\\
-0.73889562070442	0.349134557066602\\
-0.437030291029804	0.77256559702143\\
};
\addlegendentry{End Effector\\ Position}

\end{axis}
\end{tikzpicture}%

%% file: FinalFigures/2by2.tikz
\definecolor{mycolor1}{rgb}{1.00000,0.69020,0.00000}%
\definecolor{mycolor2}{rgb}{0.99608,0.38039,0.00000}%
\definecolor{mycolor3}{rgb}{0.86275,0.14902,0.49804}%
\definecolor{mycolor4}{rgb}{0.47059,0.36863,0.94118}%
\pgfplotsset{ticks=none}

\begin{tikzpicture}[font=\footnotesize]

\begin{axis}[%
axis equal,width=2.3000in,
height=1.3in,
at={(0.768in,0.496in)},
scale only axis,
unbounded coords=jump,
xmin=-1.4,
xmax=1.4,
xlabel={X (m)},
ylabel style={ align=center},
ylabel={\textbf{End Effector Position}\\[1ex]Y (m)},
% xticklabel=\empty, % Remove x-axis labels
% yticklabel=\empty, % Remove y-axis labels
axis background/.style={fill=white}
]
    
\addplot [color=mycolor1, line width=2.0pt, forget plot]
  table[row sep=crcr]{%
0.0381826754797796	0.00170187071114714\\
0.0257581550389765	0.00758973340482838\\
0.0168961215695813	0.00150780506038339\\
0.0168764515905104	0.0237278920719699\\
0.0607524067575228	0.0802034605217781\\
0.129308512765436	0.138468672397965\\
0.260880161980601	0.245934961052906\\
0.573123500451405	0.527759968159816\\
0.636734571340109	0.553601573747921\\
0.674387775147077	0.556161128483244\\
0.725949193340369	0.537549671578998\\
0.798012209447606	0.474019675003453\\
0.849854388144356	0.408142666703134\\
0.905129002703748	0.315526235198323\\
0.93145591166322	0.233147304267034\\
0.986197266994072	-0.0253482463868682\\
0.99217414857594	-0.137511689928209\\
0.986113696950923	-0.220896784024161\\
0.965506439871157	-0.292230945918082\\
0.900571492428927	-0.413130926645469\\
0.838320578523599	-0.471622900738271\\
0.803585277502006	-0.482856125642141\\
0.760594920630737	-0.493297351101367\\
0.714854962890852	-0.477673563355719\\
0.604649549245565	-0.397780740746396\\
0.382588911890872	-0.27615908404414\\
0.127238017619406	-0.130221992628667\\
0.0840179486952555	-0.0875981046394939\\
0.0347746441269533	-0.0145666105008924\\
-0.0173192069553159	0.0887591878449436\\
-0.113769472832316	0.232942531313852\\
-0.210860633135157	0.334140948803195\\
-0.282454378727279	0.385418113629734\\
-0.361578393947861	0.420134130669436\\
-0.439400430689239	0.433066960226655\\
-0.514271972188744	0.433183571911054\\
-0.576766434974204	0.415992698485094\\
-0.723213199213195	0.328310368571274\\
-0.814724821758525	0.330997997213849\\
-0.869547173192416	0.315059673855385\\
-0.922015386476759	0.273690858542408\\
-0.957283763310335	0.199359257064191\\
-0.972366839470515	0.129817656975424\\
-0.987007767903804	-0.0246081389442545\\
-0.97627859793812	-0.192760421535037\\
-0.950156726134985	-0.285620486298529\\
-0.883000515450372	-0.34052790587154\\
-0.868228754089213	-0.341267673540247\\
-0.850682861095888	-0.353277665691895\\
-0.819584253920609	-0.357953755760587\\
-0.699334221694147	-0.388267597597935\\
-0.589402609369846	-0.388738774298117\\
-0.543909254462866	-0.37348269295303\\
-0.517544341689075	-0.375365876567316\\
-0.446623451441553	-0.326893277522629\\
-0.332740416897464	-0.240626941417952\\
-0.163456275466321	-0.111943473941321\\
-0.0220191396413655	-0.019006800467152\\
};
\addplot [color=mycolor2, line join=round, line width=2.0pt, forget plot]
  table[row sep=crcr]{%
0.0475363620034801	0.00466959420600155\\
0.0900341664773756	0.0200962724697884\\
0.0932348011057049	0.0358261287913941\\
0.0376124581830519	0.0463327291382001\\
-0.0786595118530038	0.0346370849582263\\
-0.011887602744638	0.0503974876356394\\
0.077343527555678	0.0801743384743703\\
0.248185094110946	0.161582433243263\\
0.363022326474065	0.235239129967554\\
0.508195840898853	0.359304852637871\\
0.579627627144238	0.436725407788715\\
0.646846229708771	0.5676076502605\\
0.639855860574848	0.520218532869279\\
0.650914925893943	0.439353497056354\\
0.681835077355432	0.377905682910518\\
0.733443155115099	0.318867568966027\\
0.799050348892865	0.270058403414679\\
0.869437673172665	0.238589557209897\\
0.93548281221821	0.22470495018371\\
1.05220487133309	0.22648021948189\\
0.896122416323745	-0.150491984828117\\
0.86521933250654	-0.265657989365604\\
0.844481179913862	-0.410567851080509\\
0.852829323875233	-0.485636509395459\\
0.857646184231076	-0.452546152844535\\
0.845139846559513	-0.36506926179548\\
0.819307743522022	-0.331765956853765\\
0.777386526471291	-0.32094191092027\\
0.717669536805696	-0.333228497711811\\
0.55196615845639	-0.409766490047933\\
0.408406002556929	-0.4683207942371\\
0.310549618963889	-0.476042837632129\\
0.225233025869671	-0.451001694571559\\
0.159147101013083	-0.392429939028714\\
0.114450789442398	-0.302365762743449\\
0.0879446585501045	-0.191382945304053\\
0.07450092446985	-0.0142247065629608\\
0.0547513090726122	0.130227512488612\\
0.0227249009024781	0.193285873368478\\
-0.0269033025609406	0.229872635073961\\
-0.0936690465781338	0.240329553476837\\
-0.380530660290075	0.231212915718653\\
-0.488549201225389	0.249420275570537\\
-0.589397423573509	0.283138334959128\\
-0.670397510632318	0.332771711658123\\
-0.72089361747621	0.388100787522226\\
-0.739940243233748	0.44235827128373\\
-0.733307524745507	0.483276046323136\\
-0.7051866202518	0.509556774917742\\
-0.694599395775259	0.500099764560007\\
-0.700392531534624	0.467851735786422\\
-0.731457787725934	0.417376283415225\\
-0.827178900468175	0.318170935785241\\
-1.01445012780072	0.180910220973201\\
-1.10720256456638	0.133078066223208\\
-1.20222365683798	0.0928308160789149\\
-1.21907995810367	0.073002247257463\\
-1.18931993024207	0.0473005749460924\\
-0.934922753168496	-0.0891455617821069\\
-0.855505297133509	-0.148838670789497\\
-0.799346960383675	-0.214920325607926\\
-0.771185633342783	-0.278863451321414\\
-0.789160250194915	-0.331672181392029\\
-0.806238874252919	-0.359288382232318\\
-0.942286438576827	-0.46808876410262\\
-0.900281995704039	-0.46399703185558\\
-0.823688151109557	-0.445894221426622\\
-0.576607691633736	-0.378900104456978\\
-0.441315044818018	-0.328753996053939\\
-0.258698917879484	-0.25179413390688\\
-0.172342023410284	-0.19647731641843\\
-0.119583500664107	-0.147117252434386\\
-0.10046754376445	-0.109659217225688\\
-0.106962285289813	-0.0781019000127166\\
-0.156619117166204	-0.00631241073323818\\
-0.150001588449318	0.0258537369185126\\
-0.136122790478161	0.0422581060848868\\
};
\addplot [color=mycolor3, line width=2.0pt, forget plot]
  table[row sep=crcr]{%
0.0419577651104517	-0.0147192068588748\\
0.0343761372420945	-0.010213077435228\\
0.020099205403934	-0.0264262753409681\\
-0.00708445165842209	-0.0272275068100203\\
0.103730946500662	0.0585263706613315\\
0.194237079678982	0.132579463273848\\
0.29912671737797	0.220960629050211\\
0.414341377574142	0.303182545191757\\
0.546823534427639	0.408657138395645\\
0.691228092992093	0.487955364727121\\
0.739894515954567	0.500908058588721\\
0.785207675546289	0.498057808402101\\
0.823976244055067	0.483205051392485\\
0.898992478821085	0.43209877132692\\
0.927552640183601	0.407354682736363\\
0.964974594277464	0.372028348332626\\
1.02131958463554	0.265691049008851\\
1.05392179427229	0.156580872915663\\
1.06026099264745	0.0951798709304532\\
1.05701950507864	0.025300288620316\\
1.03620925632635	-0.116408963716091\\
1.00172813284444	-0.228831195041985\\
0.976817005352455	-0.291275220692994\\
0.964821710640967	-0.333828454628585\\
0.944184963423505	-0.356026309908279\\
0.926158792239984	-0.391710487958559\\
0.817687023746062	-0.490482356437228\\
0.791549280799757	-0.501093759931811\\
0.769387286577586	-0.523446359273091\\
0.720476801972193	-0.527670257278604\\
0.699017135999895	-0.539966963805475\\
0.677725461051704	-0.541498811901463\\
0.560241011366505	-0.496829153200468\\
0.475926456463301	-0.427317225307648\\
0.320296909150513	-0.271416286163949\\
0.098962115491984	-0.0949565010640054\\
0.0277221225679787	-0.0386981486435389\\
-0.0994836228947895	0.0611304557524381\\
-0.12725446865389	0.0954572450646183\\
-0.156805660527185	0.116860591999762\\
-0.2109760042115	0.180054354712945\\
-0.417873722698468	0.378184131182432\\
-0.494212408619211	0.42847260307247\\
-0.548501517174238	0.448567421064032\\
-0.607081878599291	0.466433338526297\\
-0.657815578604433	0.465015077849315\\
-0.771105329946585	0.449829507960275\\
-0.797345860135886	0.434701071991169\\
-0.838002448240547	0.38585014878709\\
-0.874205298655678	0.306342822717591\\
-0.926957643815914	0.233016593277736\\
-0.965182520922338	0.137751387973929\\
-1.00467312608475	0.0851435472152291\\
-1.01043947692045	0.0318250391585624\\
-1.01230227576194	-0.0898660215616534\\
-0.976891527466625	-0.30706449311009\\
-0.93170124004476	-0.436708080732795\\
-0.8869765021942	-0.489241156739236\\
-0.851634724047817	-0.514840064426527\\
-0.753371938957428	-0.521888346247229\\
-0.729546040991535	-0.529174628732072\\
-0.681215114420691	-0.51777056703739\\
-0.636662962808945	-0.487099279890378\\
-0.612213921899728	-0.459796119905453\\
-0.586223607865562	-0.445093789628483\\
-0.533934306041024	-0.401693425246691\\
-0.504142402235551	-0.392141094991833\\
-0.439326990156135	-0.340264504951605\\
-0.369232869826412	-0.304342277053011\\
-0.213659831461113	-0.188662378429667\\
0.0288624225788083	-0.052190463757837\\
};
\addplot [color=mycolor4, line width=2.0pt, forget plot]
  table[row sep=crcr]{%
0.021755083818947	-0.0458143383427851\\
0.0385966342348484	-0.0435538160204442\\
0.0489910615354592	-0.0139311142275313\\
0.0421780091536746	0.0231497459658964\\
0.0552299203457248	0.0267224597787236\\
0.159339826707904	0.151674615703289\\
0.236978265613364	0.229369825894717\\
0.33484951364938	0.296712216581697\\
0.371939812006825	0.316290610160189\\
0.608831004074879	0.374182694548062\\
0.818761632305006	0.383961805446174\\
0.836147545949798	0.371020373422921\\
0.848993154733135	0.331280058126845\\
0.87455828856408	0.195815528681335\\
0.939458123774244	-0.0666767578225694\\
0.949656349017169	-0.148974899430069\\
0.940041140170525	-0.221309749082734\\
0.920981777215206	-0.284557487809354\\
0.878714609328649	-0.344768022910774\\
0.818799342682201	-0.422562847662216\\
0.770668164857618	-0.442791882642584\\
0.705097801090031	-0.444881648925512\\
0.653810215397672	-0.431591618633112\\
0.574966652827176	-0.370045779233222\\
0.226165146299091	-0.116180252525954\\
0.165886474083471	-0.0580180504121243\\
0.103543469608363	-0.00881243974599555\\
0.0390014584779226	0.05673798503353\\
-0.0828806620216961	0.171478527430043\\
-0.244785546536775	0.317238882103035\\
-0.354821684634582	0.387511727495639\\
-0.46228079799847	0.433916878548819\\
-0.512141093933436	0.442853327766226\\
-0.615381778177457	0.436503117924003\\
-0.693230894332135	0.419045304923685\\
-0.771555778705271	0.39531171361133\\
-0.81728883538945	0.366984078632575\\
-0.867617067154223	0.298531031430316\\
-0.906262169565144	0.21170890580427\\
-0.96597869195699	-0.00430625528245532\\
-0.97510547979977	-0.136885023314696\\
-0.973974456103394	-0.191166426801723\\
-0.954747908908305	-0.311155339025663\\
-0.907839230356051	-0.446095387505412\\
-0.886243207122184	-0.48224415493823\\
-0.870816989286695	-0.504840038136543\\
-0.850412565347209	-0.509822476354359\\
-0.809653695824263	-0.520345259493001\\
-0.702089091914034	-0.551938433556134\\
-0.649905173211492	-0.556038326392258\\
-0.593553665554816	-0.555425732572862\\
-0.498966217384716	-0.515936771875562\\
-0.438922842929212	-0.45526217024921\\
-0.287193097119425	-0.239080065583166\\
-0.23719275032083	-0.170382611140568\\
-0.18332157864819	-0.104816113335265\\
-0.120822195960943	-0.0540459759645057\\
-0.0492215114189455	-0.00893910188861136\\
};
\addplot [color=black, line width=2.0pt, forget plot]
  table[row sep=crcr]{%
0	0\\
0.00137654807181553	-0.0181686629701405\\
0.0230740886869145	0.0223742099402824\\
0.0416605025851378	0.0415584617896105\\
0.117800361152714	0.115663421080145\\
0.305331941829929	0.287108530435952\\
0.447342337912793	0.385405153287199\\
0.551394808336153	0.439032790157762\\
0.653260097383121	0.471978000437417\\
0.747040278527992	0.477799805532746\\
0.828948457258063	0.453326579364337\\
0.896456813019593	0.398615761525255\\
0.947939798863147	0.316730234543551\\
0.982661435699631	0.213237913722782\\
1.00068935474051	0.095455919239211\\
1.00262445825665	-0.028371412662954\\
0.98923021872866	-0.149838487950308\\
0.961141130299734	-0.261062054133521\\
0.918792408616669	-0.35520351065965\\
0.86255621778029	-0.426783152210479\\
0.792947617947381	-0.471830418010733\\
0.710762146853305	-0.487938248703238\\
0.617100611626567	-0.474279394372348\\
0.513322986595544	-0.431621420372181\\
0.400987199461642	-0.362341792719657\\
0.281802373530616	-0.270410924383868\\
0.0942565097168329	-0.102372930333471\\
-0.286317082716453	0.251704608954107\\
-0.406601370134528	0.346879661463037\\
-0.520143450973644	0.419777908436293\\
-0.625193770900001	0.46607213293717\\
-0.720180885703307	0.483201903481583\\
-0.803673490131255	0.470398490883034\\
-0.874401654283331	0.428628117341929\\
-0.931313762448013	0.360476612494873\\
-0.973610400433382	0.269970288079827\\
-1.00072981494731	0.162327762762639\\
-1.01232225414462	0.0436739315078125\\
-1.00826167378013	-0.079218704084802\\
-0.98867767900925	-0.199116993963731\\
-0.953929224695956	-0.308489505907064\\
-0.904480888339244	-0.399794088440131\\
-0.840765437547723	-0.466037005379651\\
-0.803680917780924	-0.487925698009784\\
-0.763190286659182	-0.501601320346994\\
-0.719378686120824	-0.506727627037481\\
-0.672367143525724	-0.503161021113302\\
-0.569474488582068	-0.47041937633077\\
-0.456575091112927	-0.40643321633777\\
-0.336753711746484	-0.317465309668997\\
-0.0921724952541825	-0.104182053015212\\
0.0800522819851006	0.0407341324200023\\
0.184112380756906	0.114628968219838\\
};
\end{axis}

\begin{axis}[%
axis equal,width=2.3000in,
height=1.3in,
at={(0.768in,2.000in)},
scale only axis,
xmin=-1.4,
xmax=1.4,
xlabel={X (m)},
ylabel style={ align=center},
ylabel={\textbf{Quadotor Position}\\[1ex]Y (m)},
axis background/.style={fill=white},
title style={font=\bfseries},
title={10-second Period}
]
\draw[line width=2pt, black] (axis cs:0,-.54) -- (axis cs:1,-.54); % Thick black line
    \node[below] at (axis cs:0.5,-.56) {\small 1m}; % Label below the line
\addplot [color=mycolor1, dashed, line width=2.0pt, forget plot]
  table[row sep=crcr]{%
0.0144277358059096	0.0139076815807755\\
0.0342001275180485	0.048416726523291\\
0.102110318317611	0.115665232829225\\
0.294285978590949	0.277449113609598\\
0.45356381407832	0.392558212266299\\
0.552643094620594	0.448757353595909\\
0.621978872245549	0.474724723430517\\
0.6846279022453	0.482666814305367\\
0.741982272721393	0.473861706218138\\
0.795192110294445	0.447066873572485\\
0.840370085374111	0.406135582839703\\
0.88946569396573	0.332244298283198\\
0.925921511281563	0.238494281290737\\
0.94739962305352	0.129881024237114\\
0.955698857897224	-0.015281178257838\\
0.944220057046992	-0.13246437075696\\
0.91672198178475	-0.237677452743106\\
0.875170585096171	-0.326058700302328\\
0.839856492589517	-0.376683902892245\\
0.797998793170201	-0.413401673756882\\
0.751353094342196	-0.434852877602317\\
0.700215180337733	-0.440192427338086\\
0.643803730086219	-0.432190266915242\\
0.554905829889448	-0.396658285729961\\
0.45212505339815	-0.335220859613994\\
0.343606418051202	-0.253470582731052\\
0.204370977550699	-0.129998505830481\\
-0.180806206236901	0.23407655845288\\
-0.298082324786537	0.310492718414973\\
-0.413952619116085	0.363073581396155\\
-0.520285967109968	0.392338355093355\\
-0.613182883504802	0.400335200751202\\
-0.679059404322525	0.39413802292385\\
-0.734463295024884	0.37623318800648\\
-0.785377946398746	0.345630863532812\\
-0.844869162004063	0.287724431232969\\
-0.895531723926641	0.211187909074735\\
-0.933084508886883	0.120178926037163\\
-0.954749545294965	0.0247839515986706\\
-0.957959402083906	-0.0660684394578483\\
-0.944454714638752	-0.153779530537943\\
-0.915079852120566	-0.228721367480542\\
-0.871019556048351	-0.289515945266212\\
-0.816024707986352	-0.333536784760921\\
-0.765402404443649	-0.354202457416539\\
-0.7109136713814	-0.362050165756864\\
-0.64924131558285	-0.355989948987491\\
-0.580706724155846	-0.334357784468604\\
-0.4583102933127	-0.271752035218981\\
-0.0178022764724948	0.00242860492339736\\
};
\addplot [color=mycolor2, dashed, line width=2.0pt, forget plot]
  table[row sep=crcr]{%
0.0114283804731183	0.0147992474990937\\
0.0208217053138346	0.0419608833985057\\
0.0590028879546163	0.0787802052098769\\
0.244742098594152	0.218353056516412\\
0.504994166088724	0.394309091890575\\
0.59610888790059	0.437866308610201\\
0.654729624056486	0.447872946619399\\
0.708191962160228	0.439845234473452\\
0.782022066505882	0.407137671068468\\
0.854466042059743	0.357244696897028\\
0.918434216696508	0.291419338795429\\
0.953414648036024	0.233373856284371\\
0.973575741182914	0.164937963285374\\
0.979667008949455	0.0822339247216224\\
0.967702077791735	-0.0429913008723402\\
0.930381220795141	-0.213896350088193\\
0.899169923514036	-0.301677970290306\\
0.861210132014178	-0.357919853972639\\
0.825588788684197	-0.381923460600499\\
0.773463419566228	-0.397918190351525\\
0.678377538486407	-0.407240869294058\\
0.569193383419136	-0.401900947821335\\
0.478130033750905	-0.382885389926721\\
0.387013428329824	-0.344973055578185\\
0.302686665314446	-0.287436420114151\\
0.223588145415407	-0.212177440510471\\
0.0999833396717218	-0.0645512858929278\\
0.00511365728633817	0.045421742005171\\
-0.0800791639756331	0.124444764626846\\
-0.167862533787176	0.186838563668302\\
-0.365934737272201	0.301570555908855\\
-0.584806205167971	0.411036813836882\\
-0.644001200782032	0.428463300987316\\
-0.697887206632311	0.427014464182962\\
-0.749569356924099	0.406164410985671\\
-0.822351709782882	0.355363085760315\\
-0.925911847511043	0.262875685833222\\
-0.976149064080462	0.204938397029421\\
-1.00877048398672	0.14411979919538\\
-1.01848357978137	0.0767996756305496\\
-1.00429201085376	0.00481427253897437\\
-0.956966960453833	-0.116690216886598\\
-0.89339719914845	-0.274482927351829\\
-0.861903474420356	-0.34399294985184\\
-0.826091061970113	-0.374361977629038\\
-0.772243236260203	-0.386851605419378\\
-0.698298350662645	-0.381163585885399\\
-0.565206461373808	-0.346235269211038\\
-0.43093114453816	-0.291775332028932\\
-0.338456217732561	-0.240058575820105\\
-0.233016180746648	-0.159948520021405\\
0.0404872458614487	0.0817351116542837\\
};
\addplot [color=mycolor3, dashed, line width=2.0pt, forget plot]
  table[row sep=crcr]{%
0.0160786835193825	-0.0175382424353573\\
0.0293040895645134	0.00485250363818102\\
0.127876833064278	0.103861476786581\\
0.234534139729063	0.186537673222488\\
0.361088159112872	0.263580188070351\\
0.52449187543368	0.347814085540742\\
0.649251051680664	0.394370741981627\\
0.734900102313832	0.408340725954806\\
0.795325606066091	0.40390405990227\\
0.846873683319484	0.385065705295052\\
0.892530155364514	0.349880586250564\\
0.928491004425185	0.302675217827739\\
0.958380776963557	0.238152362764302\\
0.980915500637317	0.141634181378602\\
0.987561142595383	0.0278671659330911\\
0.972630490839381	-0.0923247681006375\\
0.942000522474941	-0.197114959114012\\
0.89220902600447	-0.293421449924361\\
0.844186997209699	-0.353781393623035\\
0.789902619621221	-0.399929647977358\\
0.730176855533712	-0.430907845426853\\
0.664864351349993	-0.44605449375692\\
0.5942233485375	-0.445943912357396\\
0.520898933245158	-0.429137957884239\\
0.417811356009935	-0.38445327249273\\
0.307870489273356	-0.31614955972162\\
0.201523678100742	-0.234058095449393\\
0.0339693979437226	-0.0797417319769049\\
-0.315134740607123	0.24316141086542\\
-0.419522216671405	0.312264103241489\\
-0.523835651035679	0.356500279892378\\
-0.59882086076951	0.370854258103769\\
-0.693212834474582	0.368755223265903\\
-0.757774094252548	0.351408326902961\\
-0.815485401089572	0.322101463981547\\
-0.881705398670681	0.268484119826302\\
-0.919962157039051	0.217651445572519\\
-0.957514884258158	0.137915528657215\\
-0.979397057164443	0.046057102859712\\
-0.988153461274301	-0.0524237636549792\\
-0.981708557311188	-0.150370853177702\\
-0.96180085373404	-0.240506132713723\\
-0.928826990572602	-0.312974812418703\\
-0.894585174959258	-0.358071234790027\\
-0.852190166277758	-0.389676986071941\\
-0.802238595590196	-0.410561451633799\\
-0.748759680985743	-0.416366203951323\\
-0.662088356792944	-0.405647473455393\\
-0.565637183838285	-0.374745650223845\\
-0.459028324197158	-0.325529526155512\\
-0.282685756834399	-0.221675699857287\\
-0.0935570790597404	-0.0919328385148241\\
-0.00239704713085687	-0.0241086571607422\\
};
\addplot [color=mycolor4, dashed, line width=2.0pt, forget plot]
  table[row sep=crcr]{%
-0.00199899250412949	-0.0300199220290807\\
0.00687804778688939	0.00150500407024989\\
0.0431807114143468	0.0539687584819099\\
0.132704718714264	0.147897175282355\\
0.213650767052805	0.213148487167018\\
0.306996120548366	0.269751185031278\\
0.43340710375325	0.326367459736834\\
0.535573656468955	0.357493718674191\\
0.627942340368709	0.369152707328167\\
0.69148218656044	0.364064654436598\\
0.745584507919259	0.343727314983787\\
0.792434236961171	0.31295952333014\\
0.833371367734216	0.266264617592935\\
0.873784563247897	0.188789589688908\\
0.901684576131903	0.098910168818132\\
0.915085296641585	-0.000706831082601944\\
0.911737044923696	-0.123598148231772\\
0.892313557958929	-0.217394386103026\\
0.861259094046616	-0.291353144818137\\
0.817375811444025	-0.352195352739656\\
0.777033944265677	-0.382414834145668\\
0.730883320930802	-0.399487868809773\\
0.680662318016254	-0.404558838828414\\
0.624969873383239	-0.395416642048937\\
0.540045141826634	-0.361146426717106\\
0.443809202090721	-0.30458403604416\\
0.29440647286382	-0.190374252847816\\
0.070651021271762	0.00826982489625483\\
-0.146370782016173	0.196853411444848\\
-0.285482172076418	0.292776618754915\\
-0.395079162655189	0.348169195000396\\
-0.473246097621566	0.373486333013516\\
-0.552745424963636	0.384724913750562\\
-0.646340866122642	0.378924625834067\\
-0.714040870275343	0.358802779317097\\
-0.773089671221222	0.324908777352214\\
-0.825178608332964	0.276699835703146\\
-0.884840376338379	0.192336825397982\\
-0.917149870855945	0.120045291663806\\
-0.944765552498112	0.0103534227063544\\
-0.954387951747436	-0.102227716376961\\
-0.946359634830599	-0.232903341252877\\
-0.923529348720145	-0.325883481994595\\
-0.889802471299972	-0.397527160122981\\
-0.854133796683098	-0.438312182916637\\
-0.808838128172037	-0.465581926237047\\
-0.760266617336892	-0.479564697360433\\
-0.709482815248305	-0.480134722362989\\
-0.635589651303734	-0.457589099658624\\
-0.55267813947019	-0.410857409231597\\
-0.43832254056951	-0.321936785943897\\
-0.0577981360159859	0.00972498909008612\\
};
\end{axis}

\begin{axis}[%
axis equal,width=2.3000in,
height=1.3in,
at={(3.336in,0.496in)},
scale only axis,
xmin=-1.4,
xmax=1.4,
xlabel={X (m)},
ylabel={Y (m)},
axis background/.style={fill=white},
legend style={at={(1.02,0.5)}, anchor=west, legend columns=1, legend cell align=left, align=left, draw=none}
]
\addplot [color=mycolor1, line width=2.0pt]
  table[row sep=crcr]{%
0.0124171964148823	-0.0146949914372436\\
0.000630101937868055	-0.0328624435659719\\
0.0101585664071469	-0.0552446751557003\\
0.0246621667698002	-0.0375493293449116\\
0.0262338719787449	0.0187516673150525\\
0.036459056826601	0.0174102676299979\\
0.0600653928504955	0.0132023087887427\\
0.0764923844430445	0.0316562708737129\\
0.0884812946965883	0.0771698209906015\\
0.10910372514059	0.11303679872873\\
0.143449634390179	0.165396765165131\\
0.177231303045	0.198779723925768\\
0.306641688030554	0.285954488323182\\
0.40601952470518	0.334583708963996\\
0.573214566606834	0.399115481538584\\
0.690525429625094	0.424557875010722\\
0.801207999516197	0.426198027113488\\
0.889003283407321	0.409806056268294\\
0.924512389475322	0.393044596589258\\
0.979514711131636	0.340336688480727\\
1.01439000703952	0.257682447763924\\
1.02847475039259	0.15184880136104\\
1.02423981451311	0.0244218171048538\\
1.00126537280461	-0.111511498764013\\
0.957437087412525	-0.246113116828684\\
0.898086684400445	-0.374947393867561\\
0.780786348497274	-0.508288638458612\\
0.74033732329067	-0.536526626228111\\
0.697135503408294	-0.553955741879574\\
0.650795167593934	-0.560992327534208\\
0.603643587618234	-0.555020478410163\\
0.554608398795006	-0.538907338814926\\
0.454638240723651	-0.474530389896663\\
0.350311393019254	-0.372819408304433\\
0.203734452678221	-0.181630019746064\\
0.0718444802019207	0.0249873420388267\\
-0.113140209271885	0.271059450911887\\
-0.212326102736998	0.372035917152711\\
-0.310651843657003	0.451429849410225\\
-0.411614832356775	0.505130593156662\\
-0.46091263447256	0.524231667951965\\
-0.554053682223114	0.53276581348659\\
-0.642620909801569	0.513169908247032\\
-0.680269825462483	0.492091225934185\\
-0.744819611657027	0.428607913583474\\
-0.787383123383061	0.341476294727829\\
-0.813374471139597	0.234051563622315\\
-0.825557859329018	0.115729269137329\\
-0.823612640207472	-0.0741734445104039\\
-0.80709986232451	-0.197264957741741\\
-0.783684513591332	-0.321775307996453\\
-0.72915427879789	-0.474487911123331\\
-0.6623734931719	-0.552531161720817\\
-0.636291870165155	-0.562565459048854\\
-0.608934668074475	-0.563021514220996\\
-0.547075301661817	-0.537688344648415\\
-0.474095897473052	-0.478439117322037\\
-0.348026087565721	-0.331361883281158\\
-0.136694186824626	-0.0704236696210285\\
0.1024946090217	0.149601582432065\\
};
% %\addlegendentry{ES}
%\addlegendentry{Empty Short}

\addplot [color=mycolor2, line width=2.0pt]
  table[row sep=crcr]{%
-0.0792728207020077	-0.0119389859463035\\
-0.0931946418157705	-0.0130945378623653\\
-0.026660239299868	-0.0198571464969308\\
0.0790232189692848	-0.0141494496291716\\
0.11397436512005	0.00474898058775763\\
0.115903225088862	0.0246553276940142\\
0.100482204292381	0.0357675623563327\\
0.021603810266247	0.0497867082717813\\
-0.0581696504555447	0.0397257392963284\\
-0.0428061239326023	0.0287648177533621\\
0.00175166212894906	0.0279269278805321\\
0.0523649027233593	0.040454924607574\\
0.0900185257987154	0.0660158904822108\\
0.111076235435365	0.104650195573379\\
0.118453336769986	0.13685775666034\\
0.13633708329098	0.169684228802459\\
0.174995600010294	0.209860757215423\\
0.251416934089967	0.262162005470783\\
0.379194289797463	0.323841252638956\\
0.543996867947029	0.382991000868866\\
0.666033346715033	0.406245214109046\\
0.786910621935792	0.412057405991318\\
0.844196443817179	0.405113994960087\\
0.897257599157592	0.389736874662569\\
0.988529381826237	0.337857404762049\\
1.02608971009361	0.300935411379402\\
1.05591601143197	0.257824111002799\\
1.07884329054906	0.206758953645529\\
1.09593126586402	0.0916688644071479\\
1.08459842392719	-0.0425965110023578\\
1.03417863601448	-0.182531924424257\\
0.952790286148347	-0.315578392040787\\
0.851633848383403	-0.43555736875397\\
0.796235936943311	-0.496049738473717\\
0.674356808819671	-0.555949494765459\\
0.553164990713572	-0.584979026446886\\
0.491924258824236	-0.583942827711561\\
0.388997293258134	-0.52955728799672\\
0.298997772552492	-0.443997013439706\\
0.223073582393666	-0.32884690713027\\
0.0115474518997891	0.082460855922623\\
-0.0825634723021291	0.211138454105913\\
-0.19068930018101	0.322120989254162\\
-0.314942686262716	0.407459085360093\\
-0.449263425374895	0.465867119120056\\
-0.586977664861707	0.495700841376816\\
-0.719208630716674	0.497373461111983\\
-0.835537719516986	0.475435876448983\\
-0.927712399306849	0.437495310608684\\
-0.969787018414842	0.411192232316433\\
-1.01027150447095	0.340256303376711\\
-1.02452690556648	0.251847500446736\\
-1.01181935038774	0.142062895990566\\
-0.956898921696268	-0.0545543774675281\\
-0.807316768650724	-0.52587818153288\\
-0.759409488662253	-0.62130329477625\\
-0.742617648363451	-0.63368641870194\\
-0.705170341452629	-0.638111112859672\\
-0.619590551710369	-0.524115553029679\\
-0.535001144462416	-0.405504416952698\\
-0.416199689130194	-0.261821128985086\\
-0.26455717481194	-0.0973503084927263\\
-0.09778144311853	0.0604922796623113\\
0.0711655536698252	0.186672261536105\\
0.152501519523759	0.23638818990574\\
};
%\addlegendentry{Inflated Short}

\addplot [color=mycolor3, line width=2.0pt]
  table[row sep=crcr]{%
0.0532353476092851	0.0477235285354471\\
0.0351548381480562	-0.00337800697738988\\
0.0205921048506488	-0.029522729226352\\
0.0404972432303541	-0.018228479134599\\
0.0362101595571911	0.0346316342215551\\
0.0779698014452537	-0.00952322254552485\\
0.0870034654879177	-0.00998945168847953\\
0.0879971289389152	0.0035899996675095\\
0.0913236596862665	0.0220990711379339\\
0.091637941771374	0.0862537360260174\\
0.113098918159216	0.139905545029385\\
0.167694167532952	0.198835018810288\\
0.207540196979792	0.23189761517253\\
0.371103132697477	0.32706564032268\\
0.57122534289946	0.413364222515866\\
0.754844338008391	0.464930700670978\\
0.856563121005431	0.473652004024853\\
0.946912111518434	0.455995913441412\\
0.98718681383507	0.436534545493594\\
1.02172672935304	0.40676529445581\\
1.06962531913627	0.32875167321307\\
1.0971630857519	0.221432428104404\\
1.10265306259785	0.157353676771817\\
1.08896228449374	0.0205005799621696\\
1.05345817745797	-0.124563882973207\\
0.998850214016222	-0.260577513687292\\
0.924924088031527	-0.375843905701647\\
0.844151826438243	-0.470855926993346\\
0.759582964133736	-0.542127230257729\\
0.672037781261528	-0.574381963725729\\
0.624301123433676	-0.57727674483258\\
0.576688634804109	-0.57093962916212\\
0.475547052379144	-0.529790613397692\\
0.369041159041509	-0.450740550797689\\
0.26059042999118	-0.346780063620853\\
0.0971730665169643	-0.159585835214194\\
-0.00818601618438386	-0.0405839884922414\\
-0.11965964965466	0.0903574879042275\\
-0.300035064552593	0.280789417594584\\
-0.488993046741635	0.438473716764825\\
-0.611737080885719	0.50085412410582\\
-0.72691605849007	0.527303032347254\\
-0.83342068363571	0.51940643278569\\
-0.93647964325284	0.478540075776889\\
-1.01670017207369	0.39829087818372\\
-1.05463293117939	0.346541618176373\\
-1.09929769683882	0.222880022303642\\
-1.11642430438134	0.085697745634095\\
-1.11192844106203	-0.0428485830902323\\
-1.09182638802742	-0.167832156881439\\
-1.05670261399384	-0.283201999155567\\
-1.01176047870967	-0.379920767151698\\
-0.985463394316414	-0.431583892032058\\
-0.95575352619292	-0.460817423904833\\
-0.891624622869895	-0.499686741334904\\
-0.818125989889314	-0.506766303145266\\
-0.774558820513882	-0.500905981357916\\
-0.677664864444496	-0.45551283162876\\
-0.625905490828958	-0.441897637997729\\
-0.501291443173072	-0.377697254762803\\
-0.29731337596517	-0.264790569160273\\
0.193543985592289	0.042098984123812\\
};
%\addlegendentry{Empty Long}

\addplot [color=mycolor4, line width=2.0pt]
  table[row sep=crcr]{%
-0.00759711808474894	-0.00957199227789185\\
0.0144261696083143	0.0129895699895968\\
0.041600695840897	0.0492902476542205\\
0.0622173851194239	0.0322067198658227\\
0.0316060185927554	0.0872196475511884\\
0.0326120862591639	0.0974325887388805\\
0.0350373159666391	0.1071086177206\\
0.0751210536719025	0.107161393059559\\
0.121544457246925	0.117545097943944\\
0.330527862095142	0.242294309098091\\
0.487572497277212	0.352986818726481\\
0.581704823358302	0.393090101148899\\
0.677817008796031	0.407922551995318\\
0.769776897107668	0.396696480105957\\
0.855359633386928	0.359908213894255\\
0.929793572434633	0.290571834139287\\
0.963009102286021	0.248315899309127\\
1.0028906720832	0.153728332743096\\
1.02465946334101	0.0418901123870628\\
1.02488014653012	-0.079014296037526\\
1.00881033925907	-0.200215446051074\\
0.971300725640961	-0.314072211798267\\
0.922728824904322	-0.411326163156684\\
0.8579973635051	-0.488028406551104\\
0.77672771350221	-0.539411073604361\\
0.68627863039108	-0.563242843559195\\
0.639371327299149	-0.563245956106234\\
0.533421157714273	-0.534021092117118\\
0.415161349638773	-0.471498338365638\\
0.292572143915419	-0.379564593364836\\
0.167203895951835	-0.263562009303046\\
-0.0846514891063226	0.0126376143391487\\
-0.238759275744196	0.209056826741551\\
-0.383628629701732	0.370137376605786\\
-0.477402840305516	0.446101941002005\\
-0.576104559003867	0.491871812445764\\
-0.669400027491081	0.498093185159829\\
-0.713700674363029	0.48759029816292\\
-0.799853895791778	0.441826154976935\\
-0.885197362558906	0.368285746436457\\
-0.926020444153872	0.3256292087102\\
-0.983310370605326	0.217230492314294\\
-1.0255368214562	0.0954890870615901\\
-1.04739250859555	-0.0443795730437595\\
-1.03694620163268	-0.180838882537408\\
-1.00012024820448	-0.306964318614968\\
-0.948295503409442	-0.428954913684528\\
-0.872294425841545	-0.501568652640207\\
-0.833680010734038	-0.530891563476077\\
-0.747068604023437	-0.569210727718412\\
-0.649420855919971	-0.572686139101309\\
-0.539652902269153	-0.523772083314654\\
-0.431635257904483	-0.454048194073253\\
0.0638733484352842	0.0368342748649566\\
0.154283943158247	0.121134689398083\\
};
%\addlegendentry{Inflated Long}

\addplot [color=black, line width=2.0pt]
  table[row sep=crcr]{%
0	0\\
0.00137654807181553	-0.0181686629701405\\
0.0230740886869145	0.0223742099402824\\
0.0416605025851378	0.0415584617896105\\
0.117800361152714	0.115663421080145\\
0.305331941829929	0.287108530435952\\
0.447342337912793	0.385405153287199\\
0.551394808336153	0.439032790157762\\
0.653260097383121	0.471978000437417\\
0.747040278527992	0.477799805532746\\
0.828948457258063	0.453326579364337\\
0.896456813019593	0.398615761525255\\
0.947939798863147	0.316730234543551\\
0.982661435699631	0.213237913722782\\
1.00068935474051	0.095455919239211\\
1.00262445825665	-0.028371412662954\\
0.98923021872866	-0.149838487950308\\
0.961141130299734	-0.261062054133521\\
0.918792408616669	-0.35520351065965\\
0.86255621778029	-0.426783152210479\\
0.792947617947381	-0.471830418010733\\
0.710762146853305	-0.487938248703238\\
0.617100611626567	-0.474279394372348\\
0.513322986595544	-0.431621420372181\\
0.400987199461642	-0.362341792719657\\
0.281802373530616	-0.270410924383868\\
0.0942565097168329	-0.102372930333471\\
-0.286317082716453	0.251704608954107\\
-0.406601370134528	0.346879661463037\\
-0.520143450973644	0.419777908436293\\
-0.625193770900001	0.46607213293717\\
-0.720180885703307	0.483201903481583\\
-0.803673490131255	0.470398490883034\\
-0.874401654283331	0.428628117341929\\
-0.931313762448013	0.360476612494873\\
-0.973610400433382	0.269970288079827\\
-1.00072981494731	0.162327762762639\\
-1.01232225414462	0.0436739315078125\\
-1.00826167378013	-0.079218704084802\\
-0.98867767900925	-0.199116993963731\\
-0.953929224695956	-0.308489505907064\\
-0.904480888339244	-0.399794088440131\\
-0.840765437547723	-0.466037005379651\\
-0.803680917780924	-0.487925698009784\\
-0.763190286659182	-0.501601320346994\\
-0.719378686120824	-0.506727627037481\\
-0.672367143525724	-0.503161021113302\\
-0.569474488582068	-0.47041937633077\\
-0.456575091112927	-0.40643321633777\\
-0.336753711746484	-0.317465309668997\\
-0.0921724952541825	-0.104182053015212\\
0.0800522819851006	0.0407341324200023\\
0.184112380756906	0.114628968219838\\
};
%\addlegendentry{Reference}

\end{axis}

\begin{axis}[%
axis equal,width=2.3000in,
height=1.3in,
at={(3.336in,2.000in)},
scale only axis,
xmin=-1.4,
xmax=1.4,
xlabel={X (m)},
ylabel={Y (m)},
axis background/.style={fill=white},
title style={font=\bfseries},
title={5-second Period},
legend style={at={(1.02,0.5)}, anchor=west, legend columns=1, legend cell align=left, align=left, draw=none}
]

\addlegendimage{gray, dashed, line width=2.0pt}
\addlegendentry{Quadrotor Position}
\addlegendimage{gray, line width=2.0pt}
\addlegendentry{End Effector Position}
\addlegendimage{white}
\addlegendentry{}
\addlegendimage{mycolor1, line width=2.0pt}
\addlegendentry{Empty Short}
\addlegendimage{mycolor2, line width=2.0pt}
\addlegendentry{Inflated Short}
\addlegendimage{mycolor3, line width=2.0pt}
\addlegendentry{Empty Long}
\addlegendimage{mycolor4, line width=2.0pt}
\addlegendentry{Inflated Long}
\addlegendimage{black, line width=2.0pt}
\addlegendentry{Reference}

\addplot [color=mycolor1, dashed, line width=2.0pt]
  table[row sep=crcr]{%
0.0065286025828476	-0.00898136072995093\\
0.0327708956942697	0.0451571242495548\\
0.117490241875176	0.126359505644417\\
0.180044913695335	0.18514855976298\\
0.287065044254391	0.258996986215298\\
0.390443478048369	0.309184252785136\\
0.468439946081605	0.334549772106449\\
0.547064557436516	0.347918961709277\\
0.619949916828137	0.349125013755496\\
0.69881369713782	0.336387317820294\\
0.766555762049154	0.309282591331769\\
0.828407676726455	0.264956133989825\\
0.877660092393386	0.205182886142161\\
0.911192751937041	0.132197345067744\\
0.92786926587615	0.0533092528584052\\
0.927994089448228	-0.031725084967537\\
0.911662824120785	-0.112106909051106\\
0.880203860503012	-0.185923164912994\\
0.830863597824051	-0.255194040530491\\
0.769212736586999	-0.306010681361604\\
0.694847072211204	-0.336962391865838\\
0.61074427333859	-0.345175729230979\\
0.51863174417977	-0.330453774555101\\
0.420161600763714	-0.293168645097803\\
0.323426554439928	-0.240155380437823\\
0.16469297719735	-0.127761994242632\\
-0.0487449062394573	0.0553842525638414\\
-0.291852618290604	0.263255637399997\\
-0.421179573636002	0.350783524701246\\
-0.499968273817465	0.386039034623951\\
-0.566612483181954	0.399218047890691\\
-0.631118963604481	0.389855128324563\\
-0.679346720031553	0.35962236724569\\
-0.718358237788483	0.307815677435963\\
-0.746102097690901	0.23890915486253\\
-0.763504135074183	0.153909502539273\\
-0.769941143613607	0.0636113699322879\\
-0.759864682768868	-0.0855626049930068\\
-0.729430724968926	-0.212935001757918\\
-0.696229926744009	-0.285702435102069\\
-0.653747435099769	-0.337858547853426\\
-0.601026322622332	-0.367117206302751\\
-0.53822308890567	-0.370772698548469\\
-0.469242455833134	-0.352956057207443\\
-0.389519881354164	-0.31438965081286\\
-0.264152234431197	-0.232476417401428\\
-0.0945192856572136	-0.101238823525716\\
0.0744347014321152	0.050566296276595\\
0.11671994691852	0.0908469932007691\\
};
%\addlegendentry{Empty Short}

\addplot [color=mycolor2, dashed, line width=2.0pt]
  table[row sep=crcr]{%
-0.00723287402951378	0.00145552006598515\\
0.0347875785184824	0.0130584062706152\\
0.0193612656838752	0.0483956878171036\\
0.112361805848626	0.127192476894881\\
0.192678692436383	0.190627048344183\\
0.302048133215821	0.256835566312321\\
0.405289889146398	0.300868892047721\\
0.521545753652267	0.330834193331709\\
0.606084960658958	0.338433926250456\\
0.689657336065916	0.332939217832537\\
0.772505751196054	0.311569612077249\\
0.843064242351747	0.274907535380544\\
0.900103931937826	0.220563224431469\\
0.940057624994598	0.150220748701759\\
0.960350141167364	0.0704902428761618\\
0.961242990426363	-0.0220865414990659\\
0.941970338343344	-0.115704368882915\\
0.904772670101415	-0.201423662947842\\
0.848734489909124	-0.278855793035864\\
0.776993944432896	-0.339130364494279\\
0.69879695712915	-0.374742124132294\\
0.605540421000942	-0.387927066395502\\
0.511329853077646	-0.375280981445551\\
0.409304255602246	-0.33668517712926\\
0.307781419855382	-0.275705455181127\\
0.156173637586115	-0.157730837259406\\
-0.0583864247701523	0.0407485116769148\\
-0.211176193571026	0.178802287736216\\
-0.318111090491835	0.262374444532713\\
-0.422795215922533	0.330537312701988\\
-0.52409943791427	0.380329133846461\\
-0.617503794944624	0.407260632964316\\
-0.696282779741982	0.410796160610552\\
-0.770041863244848	0.391456570897936\\
-0.826533751970832	0.350550097457957\\
-0.868767578039817	0.288496124789952\\
-0.897190169126681	0.205162994785199\\
-0.911051705930664	0.109399085609636\\
-0.911958074738583	0.00396483638068845\\
-0.901477719395348	-0.0984802939651559\\
-0.881564357822524	-0.197465471276995\\
-0.85113805424108	-0.288912044879072\\
-0.811158361612266	-0.360063718085051\\
-0.756301271358063	-0.409969771589356\\
-0.727356595243158	-0.423637743801031\\
-0.657690639876836	-0.432082242969072\\
-0.573378099292729	-0.411864912857374\\
-0.478367667983552	-0.366346313984446\\
-0.322243606721726	-0.267023879484862\\
-0.0294716528367354	-0.0522170904691416\\
0.193253301908049	0.120037375153804\\
};
%\addlegendentry{Inflated Short}

\addplot [color=mycolor3, dashed, line width=2.0pt]
  table[row sep=crcr]{%
0.0186424085663378	0.01547622941459\\
0.00936007320235321	-0.00421067753330684\\
0.0686292105830302	0.0379669737911471\\
0.143637672954927	0.117033551154134\\
0.235730308283597	0.188784513635305\\
0.32774518598331	0.241100095155957\\
0.426653991767902	0.277133948764074\\
0.531580140354152	0.290825594414345\\
0.60390833149272	0.285555292883524\\
0.672124246750795	0.268160709873419\\
0.735592833239423	0.238524540886717\\
0.787773967679787	0.199092328143061\\
0.829912697053461	0.148585696203746\\
0.857698122208619	0.091726615549515\\
0.871817948171294	0.0312555919429671\\
0.8729145652784	-0.0393033998756032\\
0.861756185105151	-0.099988734816207\\
0.837761312580771	-0.158201122208628\\
0.801177557753513	-0.207865581166679\\
0.749208530368692	-0.249469660754771\\
0.68793680567235	-0.274532482127497\\
0.620397208976641	-0.283810671213391\\
0.53727742556117	-0.278038191740151\\
0.445920046084474	-0.257162951062133\\
0.29782581899965	-0.199993593435231\\
0.0862403788146313	-0.0915665881614327\\
-0.30027177045868	0.115399215559384\\
-0.452399206230764	0.175618512813949\\
-0.550222435221485	0.200767979881887\\
-0.640154657826936	0.211559094968246\\
-0.718027272897757	0.207775930037136\\
-0.783843404203112	0.189502312502658\\
-0.840153043235362	0.156226952518655\\
-0.881706359909023	0.113083436086235\\
-0.91015459771276	0.0650868651433578\\
-0.928330298696576	0.00909658900594712\\
-0.932930181196406	-0.0485614120638929\\
-0.92467543373508	-0.10069261733707\\
-0.901642804680135	-0.153367073329751\\
-0.866966478082506	-0.193242986063449\\
-0.819248083756486	-0.223579510221669\\
-0.760112469924265	-0.242325262637011\\
-0.685308935031283	-0.250318094213162\\
-0.555220009698457	-0.241271957662827\\
-0.413801606682459	-0.21290519595428\\
-0.260294182525845	-0.164573790547311\\
-0.0531280293284657	-0.0769297810564518\\
0.105150608477322	0.00457483038097584\\
};
%\addlegendentry{Empty Long}

\addplot [color=mycolor4, dashed, line width=2.0pt]
  table[row sep=crcr]{%
0.00131882609181444	0.00241734453540976\\
0.014120034659041	0.0203488818416635\\
0.0642851091642757	0.0976027839569389\\
0.146154756394307	0.16447274604241\\
0.233618830194888	0.220026007939462\\
0.318537771534242	0.258701987320408\\
0.406954618046536	0.280703884506137\\
0.471615381679256	0.282720332915374\\
0.532502020119448	0.273074215344413\\
0.596753782795052	0.25124726314323\\
0.658603168472378	0.217863339335403\\
0.715599758239828	0.174402941290119\\
0.765071261441815	0.122426632819873\\
0.804207702104386	0.0642164496178231\\
0.832878069675193	2.66531161657868e-05\\
0.848912593966768	-0.0646348613454629\\
0.852594520871978	-0.127106887806742\\
0.84326538528411	-0.183706979832054\\
0.818469488529303	-0.234972552602701\\
0.779082540350483	-0.275769222943818\\
0.729388105487722	-0.301548360935968\\
0.663315121738296	-0.313797341757575\\
0.588125276213919	-0.308711837698763\\
0.503361200536394	-0.287211666843935\\
0.367841135735349	-0.231558286129632\\
0.207409063076615	-0.144958208471007\\
-0.271532147536355	0.137544973084314\\
-0.416069507728961	0.197156806112868\\
-0.506680280059938	0.221155680790706\\
-0.590547107632766	0.228658572336476\\
-0.667496067799224	0.218924356916865\\
-0.729136800656302	0.19456099650065\\
-0.783225527814767	0.153059583204519\\
-0.819955414274804	0.103881837013808\\
-0.845336047663371	0.0419894695639041\\
-0.857168866747801	-0.0233488646485411\\
-0.855668258718061	-0.0909941939936147\\
-0.84006037874917	-0.155273274723712\\
-0.813541234534143	-0.209200852095076\\
-0.775873731294523	-0.252392071322223\\
-0.723026818787879	-0.285406872883847\\
-0.65927188828506	-0.303020376351759\\
-0.585319439898904	-0.305279750697253\\
-0.505487615618068	-0.29290114619296\\
-0.379417286153053	-0.251969639015835\\
-0.193020867115848	-0.167810151050211\\
-0.00535241593760283	-0.0649452427951291\\
0.135140490689677	0.0225721009444179\\
};
%\addlegendentry{Inflated Long}

\end{axis}
\end{tikzpicture}%

%% file: FinalFigures/kick_xz.tikz
\definecolor{mycolor1}{rgb}{1.00000,0.69020,0.00000}%
\definecolor{mycolor2}{rgb}{0.99608,0.38039,0.00000}%
\definecolor{mycolor3}{rgb}{0.86275,0.14902,0.49804}%
\definecolor{mycolor4}{rgb}{0.47059,0.36863,0.94118}%
\begin{tikzpicture}[font=\footnotesize]

\begin{axis}[%
axis equal,
width=\columnwidth,
height=.78\columnwidth,
at={(0.758in,0.481in)},
% scale only axis,
xmin=-0.207777409311081,
xmax=1.62142528233363,
xlabel={X (m)},
ymin=0.27,
ymax=1.72,
ylabel={Z (m)},
axis background/.style={fill=white},
axis x line*=bottom,
axis y line*=left,
]
\addplot [line join=round,color=mycolor1, dashed, line width=2.0pt]
  table[row sep=crcr]{%
0.00997953957326003	1.24346931685112\\
0.0308406582406915	1.2452902788181\\
0.181538889964187	1.2414671385044\\
0.378941999336651	1.23812392396404\\
0.635032401470034	1.24641211237287\\
0.877528087623653	1.2581543808875\\
1.00903075451089	1.24619274752732\\
1.04109300631515	1.24630409739974\\
0.952472201240994	1.26026145779381\\
0.840121247262241	1.26872384416299\\
0.762260328357861	1.26512479391632\\
0.673874088174603	1.25284371269454\\
0.57814393966655	1.2498310256868\\
0.411509634574573	1.26091510191885\\
0.232344400639233	1.27197041437395\\
0.152356199973928	1.26313452191705\\
};
% \addlegendentry{QR (ES)}

\addplot [line join=round,color=mycolor1, line width=2.0pt]
  table[row sep=crcr]{%
0.0136856086038992	0.558809359702572\\
-0.0028542987269653	0.5924154509508\\
-0.0204449625607048	0.575183648426314\\
-0.0564451353952589	0.578793742183465\\
-0.0869591739072371	0.604014966278677\\
-0.101286359067336	0.650596915268799\\
-0.0535460319127614	0.697169959404435\\
0.00684591012359381	0.718164211736632\\
0.0976617959864299	0.727927642324762\\
0.195561945850658	0.701946752816857\\
0.512615402675395	0.631958888411625\\
0.70452638617629	0.605743576142755\\
0.899920569607665	0.600707073945196\\
1.08544841069208	0.619213528648207\\
1.25882198828425	0.653946211904647\\
1.39550486345969	0.722408813284896\\
1.48272127098558	0.803604208841614\\
1.53319287095888	0.874168427695757\\
1.549582326809	0.924521116896949\\
1.54448522599665	0.954981108506801\\
1.51795131972212	0.965841347300364\\
1.47266820409031	0.956653980460091\\
1.42775184819639	0.926445849842789\\
1.32853291301574	0.878409890357711\\
0.988732948970148	0.641239630846008\\
0.848167524893469	0.595613896292622\\
0.708228212040153	0.564737077689075\\
0.569810103053905	0.598747556252226\\
0.446690963294256	0.616883576000225\\
0.350420678345388	0.653537945183498\\
0.276236166986766	0.692381938134762\\
0.212955142644349	0.707299995343773\\
0.144522803365568	0.734832071394485\\
0.140983362926478	0.703146081422097\\
0.155782848864522	0.668719354718904\\
0.18337246799627	0.621724043393824\\
0.22880023021999	0.583107697312637\\
0.269662545888588	0.565643102460381\\
0.29464487943656	0.574335415624031\\
};
% \addlegendentry{EE}

\addplot [line join=round,color=mycolor2, dashed, line width=2.0pt]
  table[row sep=crcr]{%
0.00671510908158179	1.26312809938827\\
0.0117862956773935	1.26853530730757\\
0.044774344723326	1.26852936144138\\
0.406237704237921	1.22579331900748\\
0.571411387731591	1.22096137832472\\
0.744300252691227	1.23317222095335\\
0.880337989834432	1.24824875930079\\
0.958465899999746	1.24099718482574\\
0.999420975554192	1.23119908889563\\
0.901447470920166	1.23332942974045\\
0.646341353065063	1.23800539835485\\
0.450695335537911	1.21578491483673\\
0.343794593384412	1.22007117594829\\
0.306081872391072	1.23247568100801\\
0.292562556688908	1.24252472131404\\
0.264940023653301	1.24455870703096\\
0.223123137229097	1.23423362026501\\
};
% \addlegendentry{QR (IS)}

\addplot [line join=round,color=mycolor2, line width=2.0pt]
  table[row sep=crcr]{%
0.064197780382349	0.59381930287923\\
0.0728006669262982	0.598095826878246\\
0.0675675120814254	0.606147161029068\\
0.0313639456148085	0.610891165255062\\
-0.0483515615861974	0.61116197787695\\
-0.0842602885517922	0.619760860580705\\
-0.105294689888563	0.634182529449004\\
-0.112961548729229	0.658448379056409\\
-0.0932524443339955	0.685530032276202\\
-0.0587336078940717	0.699196138404907\\
0.00265674450745745	0.704025868605541\\
0.197465974447618	0.680804890685889\\
0.694446705781433	0.603964951327264\\
0.899730659714529	0.588347602436323\\
1.10621735337066	0.603695784648026\\
1.28185598050952	0.662791690815793\\
1.40783607483892	0.752318096936453\\
1.48215883993789	0.842311199161069\\
1.52061371055123	0.915215637207772\\
1.52928205908196	0.971371804384551\\
1.51447584202601	1.00681803646581\\
1.47548779333641	1.01464210395147\\
1.42072302381057	0.982765924614314\\
1.36179948395226	0.935169853503031\\
1.25476190430464	0.861030962885905\\
1.11756867940622	0.747042634851595\\
0.994178398113515	0.664964644089477\\
0.821983608284513	0.589554967464016\\
0.63641025871674	0.559112742065782\\
0.457865968834734	0.558157520408569\\
0.300506593069863	0.589482552396711\\
0.152043521315496	0.640960706253471\\
0.0685381481346725	0.684360483761507\\
-0.00913494285015415	0.737195225526336\\
-0.0586918744117415	0.782827381763071\\
-0.0877235885731587	0.828079424395165\\
-0.0769637212002949	0.819090970290886\\
0.0430057776316608	0.652445729716126\\
0.124438357802128	0.585138564792409\\
0.219684477051746	0.548201847301057\\
0.312659731521133	0.549518031771054\\
};
% \addlegendentry{EE}

\addplot [line join=round,color=mycolor3, dashed, line width=2.0pt]
  table[row sep=crcr]{%
0.0137753024383596	1.49178228793897\\
0.0129249321278171	1.50293331611107\\
0.044719287042883	1.50791572687272\\
0.297897147886927	1.49249980983251\\
0.44072302429359	1.47105807271956\\
0.510549373072882	1.46834165878598\\
0.662877644318785	1.48583373892593\\
0.844328163636759	1.50998970578301\\
0.86496245516281	1.50007729117891\\
0.881127141807722	1.47744483977647\\
0.884801556009419	1.46127357657041\\
0.874268933614248	1.45878434675094\\
0.836615070974235	1.46357898293733\\
0.724276719135571	1.46820154377314\\
0.576560375637688	1.4806530524936\\
0.49800623795848	1.46310655190337\\
0.370618655852023	1.47349276844516\\
0.0888789484753751	1.49235119458757\\
0.0205290704964722	1.48796319681001\\
};
% \addlegendentry{QR (EL)}

\addplot [line join=round,color=mycolor3, line width=2.0pt]
  table[row sep=crcr]{%
0.0384457191487668	0.499777008202354\\
0.0488577134992092	0.535609807679661\\
0.039283429430242	0.552931547665329\\
0.0132735193456597	0.56901553510422\\
-0.0265164181081763	0.580837555405905\\
-0.140171410462101	0.589829059471167\\
-0.198879708658392	0.613798400568199\\
-0.195381678114262	0.674738621941135\\
-0.130784745421019	0.738957854885446\\
-0.0298350886088274	0.794751275305921\\
0.0930463050207417	0.821581538223114\\
0.254982703561352	0.824828587858776\\
0.4247454742591	0.795397319274623\\
0.630446844098189	0.732005745659357\\
1.10610463697942	0.566428796952098\\
1.34531108635593	0.568039383819377\\
1.45051795753612	0.676740477744068\\
1.53976406104249	0.890384562470279\\
1.54439987657277	0.975419050754471\\
1.46434838829333	1.02049276461441\\
1.40070738255407	1.03141658778112\\
1.34022874354124	1.0216075129485\\
1.2936356966866	0.978813004915958\\
1.19247756717219	0.924637637830386\\
1.06914221117097	0.836399171711572\\
0.95632172912958	0.715933356939481\\
0.818783918784417	0.548444601567903\\
0.645644682886519	0.423887801520786\\
0.512807706996536	0.447128485231081\\
0.393189222058938	0.515726563589125\\
0.156879152168881	0.603044579614988\\
0.082038470528184	0.649416994375037\\
0.0195835349917943	0.667627266922853\\
-0.0372025615085461	0.659038385071997\\
-0.0738514482771717	0.638114911011049\\
-0.095101078050577	0.613407726903007\\
-0.101893382965101	0.573453206224724\\
-0.0892959138336049	0.528552296372728\\
-0.0565588005165956	0.493829561178923\\
-0.0259362072918261	0.4853559086638\\
-0.00127428392220374	0.493893076376061\\
};
% \addlegendentry{EE}

\addplot [line join=round,color=mycolor4, dashed, line width=2.0pt]
  table[row sep=crcr]{%
0.00143109265364272	1.53340806495542\\
0.000617406986112456	1.54432358076138\\
0.0206495195883933	1.54991962099765\\
0.113849493148278	1.5458378333965\\
0.531569698667692	1.51838272466338\\
0.680848001752631	1.53207138756319\\
0.808846818462194	1.54739601607166\\
0.904416977896187	1.53723695030942\\
0.929428115171101	1.52608551312033\\
0.945562759969157	1.50077286696941\\
0.916670566274194	1.50816672092649\\
0.73977277201377	1.50668064816402\\
0.615791912077496	1.51340327334889\\
0.226322985026927	1.49626560828758\\
0.0947161914799626	1.50010396497119\\
};
% \addlegendentry{QR (IL)}

\addplot [line join=round,color=mycolor4, line width=2.0pt]
  table[row sep=crcr]{%
0.0283180796294247	0.542783176021031\\
0.0373149895956975	0.546582912508032\\
0.0430371088726749	0.567989084396566\\
0.0382490136274758	0.581047237429446\\
0.0241819115916795	0.593286487737696\\
-0.00275805597544609	0.602014448795268\\
-0.039603203990376	0.606315518401959\\
-0.134280699705809	0.603743848218787\\
-0.169156849136529	0.624023174388336\\
-0.156042268892917	0.675267594976307\\
-0.11884465190067	0.718294720454457\\
-0.0430788723810378	0.756680875650455\\
0.0655244008908651	0.776826317098623\\
0.204054754684206	0.773607571634911\\
0.368083836339445	0.746143958907616\\
0.562554325373698	0.705931505074128\\
0.990021846562568	0.586413124287138\\
1.21259083207745	0.555233687350495\\
1.3897886184239	0.615389036577795\\
1.50272802127949	0.709777896666284\\
1.5776091345712	0.797404413530887\\
1.61387687185475	0.86877537670515\\
1.62142528233363	0.920975675948969\\
1.60885755647038	0.951753137789704\\
1.57706708219433	0.965726354636768\\
1.52076333856852	0.97353402444216\\
1.27167623810115	0.825953269963214\\
0.879887200594611	0.580851865145207\\
0.733739070140253	0.521080333176406\\
0.589667691660785	0.494934078573314\\
0.447336674503602	0.488144631919001\\
0.30895354217118	0.491689527430847\\
0.184215970237272	0.508380644056793\\
0.0827803850816844	0.53710120470017\\
-0.0686991130691608	0.596362602135539\\
-0.121923562928059	0.62856651032444\\
-0.19088092870243	0.687030091561293\\
-0.207777409311081	0.714688120618861\\
};
% \addlegendentry{EE}

\addplot [line join=round,color=black, line width=3.0pt, only marks, mark=o, mark options={solid, black}]
  table[row sep=crcr]{%
1.5	0.75\\
};
% \addlegendentry{EE Target}

\addplot [line join=round,color=black, densely dashed, line width=2.0pt]
  table[row sep=crcr]{%
1.02	0.4\\
1.02	1.7\\
};
% \addlegendentry{QR Limits}

\addplot [line join=round,color=black, densely dashed, line width=2.0pt]
  table[row sep=crcr]{%
-.3	1.53\\
1.6	1.53\\
};

\end{axis}
\end{tikzpicture}%

%% file: FinalFigures/kick_ux.tikz
%  http://www.mathworks.com/matlabcentral/filEnd Effectorxchange/22022-matlab2tikz-matlab2tikz
\definecolor{mycolor1}{rgb}{1.00000,0.69020,0.00000}%
\definecolor{mycolor2}{rgb}{0.99608,0.38039,0.00000}%
\definecolor{mycolor3}{rgb}{0.86275,0.14902,0.49804}%
\definecolor{mycolor4}{rgb}{0.47059,0.36863,0.94118}%
\begin{tikzpicture}[font=\footnotesize]

\begin{axis}[%
width=\columnwidth,
height=2.0in,
at={(0.758in,0.481in)},
% scale only axis,
xmin=0,
xmax=2.5,
xlabel={Time (s)},
ymin=-0.5,
ymax=2,
ylabel={X Position (m)},
axis background/.style={fill=white},
axis x line*=bottom,
axis y line*=left,
legend style={at={(.4,-0.25)}, anchor=north, legend columns=4, legend cell align=left, align=left, draw=none, fill=none}
]

\addlegendimage{white}
\addlegendentry{\textbf{ES:}}

\addplot [color=mycolor1, densely dotted, line width=2.0pt]
  table[row sep=crcr]{%
0	0.725183765626018\\
0.1	1.0978456399497\\
0.15	1.28079656819657\\
0.2	1.3819872743162\\
0.25	1.66163508164463\\
0.3	1.75541783428884\\
0.4	1.78867424128776\\
0.45	1.79268244880833\\
0.5	1.78148606716913\\
0.55	1.75048230373712\\
0.6	1.57317520768219\\
0.75	0.233535715311169\\
0.8	-0.0649943764385901\\
0.85	-0.0794421562619316\\
0.9	-0.0569356045972391\\
1	0.0210111443130492\\
1.05	0.0291083823018012\\
1.1	0.00429980108550421\\
1.15	0.0449231896991069\\
1.2	-0.0312368384594977\\
1.65	-0.0346395683600793\\
1.85	-0.0153672764474631\\
2.05	-0.000607312637714408\\
2.4	-6.24518308480759e-05\\
};
\addlegendentry{QR Position Cmd}

\addplot [color=mycolor1, dashed, line width=2.0pt]
  table[row sep=crcr]{%
0	0.00997953957325981\\
0.2	0.0109641639426576\\
0.3	0.0308406582406913\\
0.4	0.0874896831191116\\
0.45	0.125333703044845\\
0.55	0.23740019027202\\
0.7	0.464276568025598\\
0.95	0.877528087623653\\
1.05	0.975788958522615\\
1.1	1.00903075451089\\
1.15	1.03120794900931\\
1.2	1.04109300631515\\
1.25	1.03953642110937\\
1.35	1.00939641767344\\
1.5	0.91713070361038\\
1.85	0.63188252943977\\
1.95	0.532274587406806\\
2.1	0.411509634574573\\
2.4	0.194652364986153\\
2.45	0.152356199973929\\
};
\addlegendentry{QR Position}

\addplot [color=mycolor1, line width=2.0pt]
  table[row sep=crcr]{%
0	0.0136856086038994\\
0.1	0.00796885384527046\\
0.25	0.00977131050428959\\
0.35	-0.0204449625607048\\
0.45	-0.0869591739072368\\
0.5	-0.100330026080347\\
0.55	-0.101286359067335\\
0.6	-0.0535460319127612\\
0.65	0.00684591012359359\\
0.75	0.195561945850658\\
0.85	0.512615402675395\\
1.05	1.25882198828425\\
1.1	1.39550486345969\\
1.15	1.48272127098558\\
1.2	1.53319287095888\\
1.25	1.549582326809\\
1.3	1.54448522599665\\
1.35	1.51795131972212\\
1.45	1.42775184819639\\
1.6	1.11814363015031\\
1.75	0.708228212040153\\
1.85	0.446690963294256\\
1.9	0.350420678345388\\
2	0.212955142644349\\
2.05	0.171378404358458\\
2.1	0.148874596706259\\
2.2	0.140983362926478\\
2.25	0.155782848864523\\
2.3	0.18337246799627\\
2.4	0.269662545888588\\
2.45	0.29464487943656\\
};
\addlegendentry{EE Position}

\addlegendimage{white}
\addlegendentry{\textbf{IS:}}

\addplot [color=mycolor2, densely dotted, line width=2.0pt]
  table[row sep=crcr]{%
0	0.782735352122191\\
0.15	1.3269322412861\\
0.2	1.38872039270559\\
0.25	1.62473683205283\\
0.3	1.68109802617136\\
0.35	1.67669792298427\\
0.45	1.68833596482261\\
0.5	1.67767764151689\\
0.55	1.63561079614263\\
0.6	1.42987649548401\\
0.75	-0.00999661190818735\\
0.8	-0.295725973046544\\
0.85	-0.257080322475309\\
0.9	-0.152602957695175\\
1	0.131916423707683\\
1.05	0.21969959584063\\
1.1	0.221481244743515\\
1.15	0.158569952657359\\
1.2	0.0758892328538048\\
1.35	0.032927541897291\\
1.55	-0.0200859538181994\\
1.65	-0.0301755507291634\\
1.75	-0.0250014832695684\\
2	-0.000184978307437245\\
2.4	3.01717284170167e-05\\
};
\addlegendentry{QR Position Cmd}

\addplot [color=mycolor2, dashed, line width=2.0pt]
  table[row sep=crcr]{%
0	0.00671510908158179\\
0.2	0.0117862956773935\\
0.25	0.0235400525821747\\
0.35	0.0722266698833689\\
0.4	0.106917755118174\\
0.5	0.20793916747311\\
0.65	0.406237704237921\\
0.9	0.810558555836082\\
0.95	0.880337989834432\\
1.05	0.958465899999746\\
1.1	0.982866277569088\\
1.15	0.997022831164609\\
1.2	0.999420975554192\\
1.3	0.974664192513623\\
1.4	0.901447470920166\\
1.45	0.855039170842567\\
1.55	0.747850180497945\\
1.65	0.646341353065063\\
1.9	0.408866546562995\\
2	0.343794593384413\\
2.1	0.313290317889085\\
2.3	0.280678604620678\\
2.4	0.243317757695663\\
2.45	0.223123137229097\\
};
\addlegendentry{QR Position}

\addplot [color=mycolor2, line width=2.0pt]
  table[row sep=crcr]{%
0	0.064197780382349\\
0.1	0.0728006669262982\\
0.2	0.0560137651847521\\
0.3	-0.00263642197301728\\
0.4	-0.0842602885517922\\
0.45	-0.105294689888563\\
0.5	-0.112961548729229\\
0.55	-0.0932524443339955\\
0.6	-0.0587336078940717\\
0.65	0.00265674450745745\\
0.7	0.0866406302500282\\
0.75	0.197465974447618\\
0.8	0.335518741257284\\
0.85	0.533754553131412\\
0.9	0.694446705781433\\
1.05	1.28185598050952\\
1.1	1.40783607483892\\
1.15	1.48215883993789\\
1.2	1.52061371055123\\
1.25	1.52928205908196\\
1.3	1.51447584202601\\
1.35	1.47548779333641\\
1.45	1.36179948395226\\
1.5	1.25476190430464\\
1.65	0.821983608284513\\
1.75	0.457865968834735\\
1.85	0.152043521315496\\
1.95	-0.00913494285015437\\
2	-0.0586918744117417\\
2.05	-0.0837349456829357\\
2.1	-0.0877235885731587\\
2.15	-0.0769637212002952\\
2.2	-0.0530097317502936\\
2.3	0.0430057776316608\\
2.4	0.219684477051746\\
2.45	0.312659731521133\\
};
\addlegendentry{EE Position}

\addlegendimage{white}
\addlegendentry{\textbf{EL:}}

\addplot [color=mycolor3, densely dotted, line width=2.0pt]
  table[row sep=crcr]{%
0	0.608948508795091\\
0.15	1.44459157195298\\
0.2	1.60124444928673\\
0.25	1.88550850724812\\
0.3	1.92397715757609\\
0.35	1.83055161644165\\
0.4	1.68778980023055\\
0.6	0.98936058719242\\
0.7	0.230286397454926\\
0.75	-0.00402222850092659\\
0.8	-0.188173904954876\\
0.85	-0.0768408147282429\\
0.9	-0.0238371702382398\\
0.95	-0.090518306216385\\
1	-0.346296804206592\\
1.05	-0.235268198745056\\
1.1	0.275180353952013\\
1.15	0.0752678110337559\\
1.2	-0.00780561867102714\\
1.3	-0.0129915948688248\\
1.6	-0.00636679522857264\\
2.1	0.00148401361605055\\
2.4	3.29066814384049e-05\\
};
\addlegendentry{QR Position Cmd}

\addplot [color=mycolor3, dashed, line width=2.0pt]
  table[row sep=crcr]{%
0	0.0137753024383596\\
0.25	0.0168551936663865\\
0.35	0.044719287042883\\
0.4	0.0714728031720124\\
0.5	0.164022396237175\\
0.75	0.510549373072882\\
0.85	0.662877644318785\\
0.9	0.727382010291048\\
1	0.816508156142043\\
1.1	0.86496245516281\\
1.15	0.881127141807722\\
1.2	0.884801556009419\\
1.3	0.856257144676041\\
1.4	0.812621002476942\\
1.5	0.754740545299453\\
1.6	0.68760489280279\\
1.8	0.539957136790731\\
2.35	0.0888789484753749\\
2.45	0.020529070496472\\
};
\addlegendentry{QR Position}

\addplot [color=mycolor3, line width=2.0pt]
  table[row sep=crcr]{%
0	0.038445719148767\\
0.35	0.0392834294302422\\
0.4	0.0132735193456597\\
0.45	-0.0265164181081765\\
0.55	-0.140171410462101\\
0.6	-0.198879708658392\\
0.65	-0.195381678114262\\
0.7	-0.130784745421019\\
0.75	-0.0298350886088272\\
0.8	0.0930463050207417\\
0.9	0.4247454742591\\
1	0.856854852803879\\
1.1	1.34531108635593\\
1.15	1.45051795753612\\
1.2	1.50218929751358\\
1.25	1.53976406104249\\
1.3	1.54439987657277\\
1.35	1.46434838829333\\
1.45	1.34022874354124\\
1.5	1.2936356966866\\
1.55	1.19247756717219\\
1.7	0.818783918784418\\
1.75	0.645644682886519\\
1.9	0.270151735753108\\
1.95	0.156879152168881\\
2.05	0.0195835349917943\\
2.1	-0.0372025615085461\\
2.15	-0.0738514482771717\\
2.2	-0.0951010780505772\\
2.25	-0.101893382965101\\
2.3	-0.0892959138336047\\
2.45	-0.00127428392220397\\
};
\addlegendentry{EE Position}

\addlegendimage{white}
\addlegendentry{\textbf{IL:}}

\addplot [color=mycolor4, densely dotted, line width=2.0pt]
  table[row sep=crcr]{%
0	0.869694132571369\\
0.15	1.45949004188803\\
0.2	1.5602702324377\\
0.25	1.82235294105321\\
0.3	1.879680959523\\
0.35	1.84021440186762\\
0.4	1.77544813388442\\
0.45	1.68609326317576\\
0.55	1.4629545909951\\
0.6	1.22677885691964\\
0.7	0.301520378415113\\
0.75	-0.0650324622421392\\
0.8	-0.290870669166743\\
0.85	-0.203225685290409\\
0.9	-0.144139908638074\\
0.95	-0.06302718761774\\
1	-0.0236438180497283\\
1.05	-0.0314202024440293\\
1.15	0.139818495037274\\
1.2	0.0171217722716159\\
1.5	0.0250513074075642\\
1.75	-0.0031187163928128\\
1.9	-0.00783242493112901\\
2.4	6.90191218164493e-05\\
};
\addlegendentry{QR Position Cmd}

\addplot [color=mycolor4, dashed, line width=2.0pt]
  table[row sep=crcr]{%
0	0.00143109265364272\\
0.25	0.0070734574888518\\
0.35	0.040271353716169\\
0.4	0.0715422425651662\\
0.5	0.166217230546453\\
0.6	0.296366563198802\\
0.9	0.75513835295299\\
1.05	0.904416977896187\\
1.1	0.929428115171101\\
1.15	0.943649124303459\\
1.25	0.940282556861749\\
1.35	0.916670566274194\\
1.45	0.87370931098928\\
1.55	0.812323080838025\\
1.7	0.693804014493107\\
1.9	0.529251966907078\\
2.1	0.342361111793346\\
2.2	0.257638610473919\\
2.35	0.156029111035601\\
2.45	0.0947161914799626\\
};
\addlegendentry{QR Position}

\addplot [color=mycolor4, line width=2.0pt]
  table[row sep=crcr]{%
0	0.0283180796294249\\
0.3	0.0382490136274756\\
0.35	0.0241819115916795\\
0.4	-0.00275805597544609\\
0.5	-0.0853966525244436\\
0.55	-0.134280699705809\\
0.6	-0.169156849136529\\
0.65	-0.156042268892917\\
0.7	-0.11884465190067\\
0.75	-0.0430788723810376\\
0.8	0.0655244008908653\\
0.85	0.204054754684206\\
0.9	0.368083836339445\\
1	0.768212838458898\\
1.1	1.21259083207745\\
1.15	1.3897886184239\\
1.2	1.50272802127949\\
1.25	1.5776091345712\\
1.3	1.61387687185475\\
1.35	1.62142528233363\\
1.4	1.60885755647038\\
1.45	1.57706708219433\\
1.55	1.46229853289026\\
1.6	1.37566809413143\\
1.7	1.14969810907628\\
1.8	0.879887200594611\\
2	0.30895354217118\\
2.05	0.184215970237272\\
2.15	-0.000401553248626563\\
2.2	-0.0686991130691608\\
2.25	-0.12192356292806\\
2.3	-0.162138400843332\\
2.35	-0.19088092870243\\
2.4	-0.206356357272973\\
2.45	-0.207777409311081\\
};
\addlegendentry{EE Position}

\addlegendimage{white}
\addlegendentry{}
\addlegendimage{white}
\addlegendentry{}

\addplot [color=black, densely dashed, line width=2pt]
  table[row sep=crcr]{%
0	1.02\\
2.5	1.02\\
};
\addlegendentry{QR Position\\Limits}

\addplot [color=black, line width=3.0pt, only marks, mark=o, mark options={solid, black}]
  table[row sep=crcr]{%
1.25	1.5\\
};
\addlegendentry{EE Target\\ Position}

\end{axis}
\end{tikzpicture}%

%% file: FinalFigures/growing_xyz.tikz
\definecolor{mycolor1}{rgb}{0.39216,0.56078,1.00000}%
\begin{tikzpicture}[font=\footnotesize]

\begin{axis}[%
width=\columnwidth,
height=1.5in,
at={(0.758in,2.903in)},
%scale only axis,
xmin=0,
xmax=20,
ymin=-1.06819529483739,
ymax=1.12890762534264,
ylabel={X (m)},
axis background/.style={fill=white},
axis x line*=bottom,
axis y line*=left,
xmajorgrids,
ymajorgrids,
legend style={at={(.5,1)}, anchor=south, legend cell align=left, align=left, draw=none, legend columns=3}
]
\addplot [color=black, line width=2.0pt]
  table[row sep=crcr]{%
0.199999999999999	0\\
0.699999999999999	0.00213377274598159\\
1.95	0.0525641135341672\\
2.85	0.126575398947356\\
2.95	0.156093679587599\\
3.05	0.201329976873108\\
3.15	0.265658383366201\\
3.3	0.396924918778428\\
3.7	0.789568564791839\\
3.85	0.896382784962789\\
3.95	0.948254567494452\\
4.05	0.983626774280808\\
4.15	1.00233094910189\\
4.25	1.00473508488846\\
4.35	0.99148594250871\\
4.45	0.963232110235008\\
4.6	0.893883622340482\\
4.75	0.793873034577992\\
4.9	0.666051674206425\\
5.05	0.514482941690634\\
5.25	0.284029759445009\\
6	-0.621062177621372\\
6.15	-0.760012708518541\\
6.3	-0.872086405791734\\
6.45	-0.95343273744087\\
6.55	-0.989104392544835\\
6.65	-1.00915919555777\\
6.75	-1.01317721208685\\
6.85	-1.00100904553867\\
6.95	-0.972829013788509\\
7.05	-0.929146127341067\\
7.2	-0.836346409759429\\
7.35	-0.713942722007655\\
7.5	-0.566093633561575\\
7.7	-0.338477402863163\\
8.1	0.163412652189916\\
8.35	0.464325204436008\\
8.55	0.674052856583618\\
8.7	0.804421330210065\\
8.85	0.90655054760165\\
9	0.976642849182408\\
9.1	1.00414910731704\\
9.2	1.01567789473691\\
9.3	1.01106151722378\\
9.4	0.990427979851383\\
9.5	0.954173411875228\\
9.65	0.871976425538787\\
9.8	0.759314845764496\\
9.95	0.620469724909018\\
10.15	0.403350138160548\\
10.4	0.0983360777578994\\
10.85	-0.458437912830348\\
11.05	-0.669288506443493\\
11.2	-0.800055753669465\\
11.35	-0.902255485956481\\
11.5	-0.972511562759276\\
11.6	-1.00039669776678\\
11.7	-1.0125172702222\\
11.8	-1.0085685379003\\
11.9	-0.988509767290541\\
12	-0.952628214252233\\
12.15	-0.870587824376301\\
12.3	-0.757643048250429\\
12.45	-0.617855156846531\\
12.65	-0.398062225705409\\
12.95	-0.0264123478673355\\
13.3	0.406643927184959\\
13.5	0.624861126018949\\
13.65	0.764108470401254\\
13.8	0.876934431887921\\
13.95	0.958941580890453\\
14.05	0.994672080182234\\
14.15	1.01439803470724\\
14.25	1.01781469479542\\
14.35	1.00498041371768\\
14.45	0.976247386206751\\
14.55	0.932195875029819\\
14.7	0.839150129264265\\
14.85	0.716899618865906\\
15	0.569936475178743\\
15.2	0.344748294279082\\
15.5	-0.0284537884044589\\
15.85	-0.458737437535493\\
16.05	-0.670527590965868\\
16.2	-0.801507128561987\\
16.35	-0.903408482780847\\
16.5	-0.973306587960188\\
16.6	-1.00111211356434\\
16.7	-1.01327430785953\\
16.8	-1.00940527642327\\
16.9	-0.989393042385402\\
17	-0.953513983659558\\
17.15	-0.871496912791155\\
17.3	-0.75874809041386\\
17.45	-0.61929792995419\\
17.65	-0.399593766786449\\
17.95	-0.0263370164294123\\
18.3	0.406729825395239\\
18.5	0.623391081707368\\
18.65	0.761884923971099\\
18.8	0.874886558007876\\
18.95	0.957716815362037\\
19.05	0.993943881579369\\
19.15	1.01386627721808\\
19.25	1.01716164638173\\
19.35	1.00399779372296\\
19.45	0.974671796994137\\
19.55	0.929376590490737\\
19.7	0.832640421569426\\
19.9	0.664610739447507\\
20.05	0.539879839410428\\
20.15	0.473844467782342\\
};
\addlegendentry{End Effector\\ Reference}

\addplot [color=mycolor1, line width=2.0pt]
  table[row sep=crcr]{%
0.199999999999999	0.0476326390257569\\
0.300000000000001	0.0100631492779577\\
0.399999999999999	-0.0188486146634119\\
0.449999999999999	-0.0248718682634852\\
0.5	-0.0588284259482776\\
0.600000000000001	-0.0779640543592883\\
0.699999999999999	-0.0794111743695929\\
0.800000000000001	-0.0630122185935633\\
1.25	0.145036121944049\\
1.35	0.167304899487757\\
1.45	0.167343606500282\\
1.55	0.146464719579903\\
1.6	0.130779174181679\\
1.9	-0.0188836969198363\\
2	-0.0466444989160202\\
2.1	-0.050165558696964\\
2.2	-0.0221393999217234\\
2.3	0.0306532879981489\\
2.6	0.216715967774114\\
2.7	0.247115522702071\\
2.8	0.256574051899189\\
2.9	0.249113395398297\\
3.1	0.218729225328641\\
3.2	0.221024198287392\\
3.3	0.248717367421534\\
3.4	0.333432327483802\\
3.55	0.512821514399629\\
3.6	0.598143733049739\\
3.65	0.659847311436671\\
3.75	0.819172628464472\\
3.85	0.949471769797416\\
3.95	1.0494888642353\\
4.05	1.11352838476978\\
4.1	1.12890762534264\\
4.2	1.12839860187785\\
4.3	1.10141406676376\\
4.45	1.01415717246203\\
4.65	0.853944004829831\\
4.7	0.821279993383921\\
4.85	0.69656058856707\\
4.9	0.670005950322967\\
5.05	0.54280091377354\\
5.2	0.4145687261401\\
5.3	0.316965093339689\\
5.35	0.280481923550091\\
5.4	0.193414887205478\\
5.55	-0.00987567414939861\\
6.05	-0.719841955942513\\
6.2	-0.878514580587954\\
6.35	-0.983768755448203\\
6.45	-1.02325260312325\\
6.55	-1.04012410034674\\
6.65	-1.04773634270401\\
6.8	-1.03767120753302\\
7.1	-0.962217470489833\\
7.25	-0.901876102484607\\
7.4	-0.821267984898128\\
7.55	-0.700544388070711\\
7.7	-0.544729626620981\\
7.8	-0.426593008299722\\
8	-0.157067374913893\\
8.15	0.0782033701153928\\
8.2	0.10631034175805\\
8.25	0.243318519331655\\
8.45	0.556601954400772\\
8.55	0.682050887715217\\
8.6	0.718253612952346\\
8.65	0.78146332941769\\
8.75	0.859411466500262\\
8.85	0.914951434759555\\
8.95	0.954502394624782\\
9.25	0.996068875930845\\
9.35	0.981821260554195\\
9.55	0.938400214085203\\
9.8	0.846997331506174\\
9.85	0.824735818020152\\
10.1	0.611568979970755\\
10.25	0.436558062499415\\
10.8	-0.458061984247554\\
10.9	-0.578546676132582\\
11.05	-0.698062505958699\\
11.15	-0.757354631398577\\
11.3	-0.810346618195389\\
12.1	-1.03206249586984\\
12.2	-1.02018335636066\\
12.3	-0.993173128381297\\
12.35	-0.970536207232922\\
12.45	-0.873713763986519\\
12.55	-0.771644058620634\\
12.7	-0.579381895241397\\
13.05	0.052244601496092\\
13.1	0.125503616092885\\
13.2	0.296146835180974\\
13.3	0.425368740100357\\
13.4	0.520892406756573\\
13.5	0.602543795018104\\
13.65	0.698152594654605\\
13.9	0.807792345627814\\
14	0.849440802423185\\
14.3	0.998717047674447\\
14.35	1.00255617197378\\
14.4	1.03510388934754\\
14.5	1.05425641720648\\
14.6	1.04144999380727\\
14.65	1.03076053486501\\
14.75	0.99606768542872\\
15.05	0.653384983532753\\
15.1	0.579085326854688\\
15.15	0.53119613326518\\
15.2	0.426599842065755\\
15.45	0.0333985311119562\\
15.6	-0.183772541033061\\
15.7	-0.293206990422693\\
15.85	-0.417579915759017\\
16.15	-0.618523092992856\\
16.45	-0.818567239673147\\
16.7	-0.986401956233916\\
16.75	-1.01408842761196\\
16.8	-1.02296323488114\\
16.85	-1.05342475999032\\
16.95	-1.06819529483739\\
17.05	-1.05446327060401\\
17.15	-1.01038713552147\\
17.2	-0.984211204223435\\
17.4	-0.804611174547137\\
17.6	-0.587669028202189\\
17.8	-0.323395191861781\\
18	-0.0481105341029711\\
18.05	-0.00826071416092589\\
18.1	0.114454626799429\\
18.25	0.307769662174465\\
18.35	0.414623782132455\\
18.55	0.583226968928479\\
18.8	0.754645389158334\\
19.05	0.896944718253639\\
19.3	0.988168159216123\\
19.45	0.994930967544985\\
19.55	0.982251069323354\\
19.65	0.953469883461953\\
19.8	0.855787398304269\\
19.95	0.715922486121176\\
20.15	0.491813128418041\\
};
\addlegendentry{End Effector\\ Position}

\addplot [color=mycolor1, dashed, line width=2.0pt]
  table[row sep=crcr]{%
0.199999999999999	0.00915453856788062\\
0.5	0.00130823036906591\\
0.75	0.0177423421382912\\
1.05	0.0469293576752072\\
1.3	0.0564013303144613\\
2	0.0507845651422585\\
2.2	0.0759845452665147\\
2.75	0.166635726746019\\
2.95	0.224020262297564\\
3.1	0.292518861969402\\
3.25	0.386263150210755\\
3.4	0.493769715851656\\
3.75	0.767017356896176\\
3.9	0.863135762271895\\
4	0.908788021229739\\
4.1	0.937952000863799\\
4.2	0.949741205894668\\
4.3	0.944868743460564\\
4.4	0.925130180158966\\
4.5	0.892298995631123\\
4.6	0.846733863646342\\
4.75	0.752242797595382\\
4.9	0.635848707226412\\
5.1	0.444851524798629\\
5.2	0.340556140653206\\
5.9	-0.486311621223148\\
6.05	-0.636611180157338\\
6.3	-0.819094654624433\\
6.4	-0.869387294126888\\
6.55	-0.92342473931652\\
6.7	-0.947242115622497\\
6.85	-0.945173793638102\\
7	-0.916404283184733\\
7.1	-0.881496889289085\\
7.2	-0.83359126100401\\
7.35	-0.73933294696112\\
7.45	-0.657494689074614\\
7.6	-0.518979992151348\\
7.7	-0.410889831410408\\
7.8	-0.2911658417563\\
8	-0.0439403747116742\\
8.15	0.152554663885976\\
8.3	0.3271266649996\\
8.4	0.442033310120362\\
8.65	0.679145551920367\\
8.9	0.842533232560694\\
9	0.887231486989279\\
9.15	0.929156224728587\\
9.3	0.939060909917789\\
9.4	0.927887900727125\\
9.5	0.903146978323804\\
9.65	0.834241807445192\\
9.8	0.73465400374781\\
9.9	0.652285552963694\\
10.1	0.448596603599448\\
10.8	-0.367179535550754\\
10.9	-0.462387718993071\\
11	-0.553903729016213\\
11.15	-0.66757438766551\\
11.25	-0.736015792772022\\
11.5	-0.858075385016409\\
11.6	-0.890948783344687\\
11.75	-0.920432861168138\\
11.85	-0.922576688470848\\
11.95	-0.911699040705127\\
12.1	-0.869350611765473\\
12.25	-0.792097481440702\\
12.35	-0.723572883498683\\
12.45	-0.639643834172603\\
12.6	-0.492388865939546\\
12.75	-0.327637684613698\\
12.95	-0.0843550165505071\\
13.4	0.410541472976966\\
13.55	0.550788314122666\\
13.85	0.775114539399315\\
14	0.851314166404826\\
14.15	0.903358970310318\\
14.25	0.920826436904008\\
14.35	0.922287022549593\\
14.45	0.907046116530683\\
14.55	0.874482660538629\\
14.65	0.827769946591268\\
14.8	0.72498751342912\\
15	0.537534716182151\\
15.25	0.268343315379077\\
15.45	0.0358833718488825\\
15.55	-0.0741045581032331\\
15.65	-0.185284682475054\\
16.1	-0.605437661416843\\
16.3	-0.749170362398306\\
16.45	-0.831733525794643\\
16.6	-0.885295759433987\\
16.75	-0.906448087525153\\
16.85	-0.903186743719129\\
17	-0.872485737312545\\
17.1	-0.834891775322664\\
17.3	-0.71543607193583\\
17.45	-0.596294813189839\\
17.6	-0.45035205435029\\
17.8	-0.239424931888848\\
17.9	-0.128273766708638\\
17.95	-0.0784274515046981\\
18.05	0.0364571821388679\\
18.15	0.150583077051269\\
18.6	0.575511377219652\\
18.7	0.651946122764517\\
18.9	0.775130996828132\\
19.1	0.852822389634316\\
19.25	0.874362789118759\\
19.35	0.870873327737417\\
19.5	0.837443274400826\\
19.6	0.802313858236033\\
19.8	0.699870607975889\\
20.15	0.508022829303211\\
};
\addlegendentry{Quadrotor\\ Position}

\end{axis}

\begin{axis}[%
width=\columnwidth,
height=1.5in,
at={(0.758in,1.692in)},
%scale only axis,
xmin=0,
xmax=20,
ymin=-0.661177087485835,
ymax=0.610253229920115,
ylabel={Y (m)},
axis background/.style={fill=white},
axis x line*=bottom,
axis y line*=left,
xmajorgrids,
ymajorgrids
]
\addplot [color=black, line width=2.0pt, forget plot]
  table[row sep=crcr]{%
0.199999999999999	0\\
0.399999999999999	-0.00496838864632565\\
0.649999999999999	-0.0175724535333508\\
0.800000000000001	-0.0118130228193571\\
1.4	0.0335914761621616\\
1.85	0.0452251959039387\\
2.15	0.0732416158660314\\
2.4	0.0924644638857011\\
2.75	0.108401110410124\\
2.85	0.125186178340531\\
2.95	0.154574530778277\\
3.05	0.198223733385475\\
3.15	0.255100315340758\\
3.4	0.414023614895108\\
3.5	0.458764311929954\\
3.55	0.472512435071984\\
3.6	0.479448409212811\\
3.65	0.479035945255227\\
3.7	0.470955313130482\\
3.75	0.455101179709651\\
3.8	0.431574977741597\\
3.85	0.400672926968006\\
3.95	0.318799038585254\\
4.05	0.215042056810308\\
4.2	0.0346957235463208\\
4.4	-0.207916013565761\\
4.5	-0.312099418651076\\
4.6	-0.396009236894553\\
4.7	-0.454834399668137\\
4.75	-0.473785513897823\\
4.8	-0.485408356408005\\
4.85	-0.489566905662926\\
4.9	-0.486227432446981\\
4.95	-0.475457419126762\\
5	-0.457424543747859\\
5.05	-0.432395598468478\\
5.15	-0.362903091427704\\
5.25	-0.271023423354535\\
5.35	-0.162197008143043\\
5.75	0.301150278359945\\
5.85	0.386963844635872\\
5.95	0.448281811000687\\
6	0.468491186735942\\
6.05	0.4812799971305\\
6.1	0.486451440621714\\
6.15	0.483925641274343\\
6.2	0.47373985719473\\
6.25	0.456047763973174\\
6.3	0.431117835773311\\
6.4	0.361175048556017\\
6.5	0.268218503402867\\
6.6	0.158058785094145\\
6.95	-0.257329950062378\\
7.05	-0.353194967051014\\
7.15	-0.426809465567231\\
7.2	-0.453791462010404\\
7.25	-0.473642786316152\\
7.3	-0.486069749446145\\
7.35	-0.490896099414091\\
7.4	-0.488063974789661\\
7.45	-0.477632844363949\\
7.5	-0.459776667136467\\
7.55	-0.434779634442844\\
7.65	-0.365018716718343\\
7.75	-0.27261327307875\\
7.85	-0.163191785225315\\
8.25	0.302169156786888\\
8.35	0.388583535366823\\
8.45	0.450598761504036\\
8.5	0.471178126228043\\
8.55	0.484336726090664\\
8.6	0.489864079186169\\
8.65	0.487665600020321\\
8.7	0.477763997287454\\
8.75	0.460299736985569\\
8.8	0.435530483581566\\
8.9	0.365681205477049\\
9	0.272510503962142\\
9.1	0.161880668772127\\
9.45	-0.255552607727981\\
9.55	-0.351710048009899\\
9.65	-0.425453624246114\\
9.7	-0.452462387989719\\
9.75	-0.472332257197944\\
9.8	-0.484782332931925\\
9.85	-0.489647248241099\\
9.9	-0.486876814284464\\
9.95	-0.476534194192997\\
10	-0.45879293354535\\
10.05	-0.433933154460018\\
10.15	-0.364484568333111\\
10.25	-0.27238236950447\\
10.35	-0.163191282873829\\
10.75	0.301899044670726\\
10.85	0.388109997227733\\
10.95	0.449764151453383\\
11	0.47010650837376\\
11.05	0.482998207685998\\
11.1	0.488238164114076\\
11.15	0.485743326224533\\
11.2	0.475549150402916\\
11.25	0.457808962152658\\
11.3	0.432792257544563\\
11.4	0.362569864655054\\
11.5	0.269219140261882\\
11.6	0.158596739282054\\
11.95	-0.257996984219027\\
12.05	-0.353843480554421\\
12.15	-0.427267778326122\\
12.2	-0.454114906498411\\
12.25	-0.473824100279913\\
12.3	-0.486117608582255\\
12.35	-0.490834096027978\\
12.4	-0.487927663679208\\
12.45	-0.477465141892473\\
12.5	-0.459622163623635\\
12.55	-0.434678556450923\\
12.65	-0.365100893752654\\
12.75	-0.272873407505788\\
12.85	-0.163494912722754\\
13.25	0.303400039953939\\
13.35	0.390301390842541\\
13.45	0.452722974131703\\
13.5	0.473461369263617\\
13.55	0.48674168705535\\
13.6	0.492347594643444\\
13.65	0.490179386968908\\
13.7	0.480256073077147\\
13.75	0.462716440295495\\
13.8	0.437818872703854\\
13.9	0.367569120074208\\
14	0.273850032919047\\
14.1	0.162588834497083\\
14.45	-0.256904636002094\\
14.55	-0.353438567671169\\
14.65	-0.427426214295625\\
14.7	-0.454512639874878\\
14.75	-0.474436946630281\\
14.8	-0.486924064362615\\
14.85	-0.491813951865222\\
14.9	-0.489059909481895\\
14.95	-0.47872549664034\\
15	-0.460980653714135\\
15.05	-0.436097604071101\\
15.15	-0.36649449817364\\
15.25	-0.274003542459024\\
15.35	-0.164126875633009\\
15.75	0.304472575135797\\
15.85	0.391039447621132\\
15.95	0.452805311518148\\
16	0.473138217083527\\
16.05	0.485990179081277\\
16.1	0.491167651890965\\
16.15	0.488592808853191\\
16.2	0.478304249523116\\
16.25	0.460456892130455\\
16.3	0.435320912268462\\
16.4	0.364824865857617\\
16.5	0.271154802600478\\
16.6	0.160162773952774\\
16.95	-0.258250428093877\\
17.05	-0.354655345761472\\
17.15	-0.428551785091301\\
17.2	-0.455595833248893\\
17.25	-0.475477603693552\\
17.3	-0.487921949464994\\
17.35	-0.492768683581907\\
17.4	-0.489969361684459\\
17.45	-0.479582859503367\\
17.5	-0.461770874545671\\
17.55	-0.436794373480591\\
17.65	-0.366877304813478\\
17.75	-0.273853903549085\\
17.85	-0.163242285846717\\
18.25	0.309888509644896\\
18.35	0.398411237469666\\
18.45	0.462522991461089\\
18.5	0.484076211026835\\
18.55	0.49808988869804\\
18.6	0.504285718059037\\
18.65	0.502495936207762\\
18.7	0.492673625036669\\
18.75	0.474903382474029\\
18.8	0.449410890632169\\
18.9	0.376899891235617\\
19	0.279852579127176\\
19.1	0.164938788644836\\
19.45	-0.259053874701866\\
19.55	-0.351726588310392\\
19.65	-0.418914660677501\\
19.7	-0.441915610844458\\
19.75	-0.457900062823832\\
19.85	-0.471321201557839\\
20	-0.465060139274211\\
20.1	-0.463442044091941\\
20.15	-0.467286087001149\\
};
\addplot [color=mycolor1, line width=2.0pt, forget plot]
  table[row sep=crcr]{%
0.199999999999999	-0.0208510476545172\\
0.350000000000001	-0.0623808461916155\\
0.449999999999999	-0.0719632794614675\\
0.5	-0.090446482555997\\
0.600000000000001	-0.0941688136486825\\
0.699999999999999	-0.0804846782765409\\
0.850000000000001	-0.0352753260971248\\
1.2	0.0965418215364728\\
1.3	0.114806273370299\\
1.4	0.117863310368662\\
1.5	0.104839430930276\\
1.65	0.0670340310032991\\
1.85	0.00723227581780606\\
1.95	-0.0173158917149934\\
2.05	-0.0313082924364885\\
2.1	-0.0362358028878695\\
2.2	-0.030503055072522\\
2.3	-0.0128029333105815\\
2.45	0.028102599437446\\
2.55	0.0723528019618449\\
2.8	0.174454139136895\\
2.9	0.201021243771134\\
3	0.219159288183477\\
3.3	0.255289183623677\\
3.45	0.282416278721762\\
3.65	0.31080578704535\\
3.75	0.325002247574879\\
3.9	0.331569153204946\\
4	0.319980400939617\\
4.05	0.306066359022171\\
4.15	0.259580333611613\\
4.2	0.226216772764154\\
4.3	0.138909479577812\\
4.35	0.0704499619749264\\
4.4	0.0149028518363465\\
4.45	-0.0582516146055134\\
4.5	-0.151618735388542\\
4.65	-0.38739323927701\\
4.7	-0.445773740404526\\
4.75	-0.51779693800226\\
4.8	-0.572848697002179\\
4.85	-0.603849865838345\\
4.95	-0.635321063208707\\
5	-0.629460909444663\\
5.05	-0.610413602459762\\
5.1	-0.57871098694979\\
5.15	-0.5348541909059\\
5.2	-0.479436680993583\\
5.3	-0.348444853016147\\
5.35	-0.294635225231783\\
5.4	-0.167229064622671\\
5.45	-0.0804808544957325\\
5.5	-0.00712652833674809\\
5.55	0.0901865114520888\\
5.65	0.23284052786277\\
5.8	0.388140397136205\\
5.85	0.420297565284741\\
5.9	0.442859076386831\\
5.95	0.456252038478851\\
6	0.461597176960659\\
6.1	0.449494615456189\\
6.2	0.416830583377887\\
6.25	0.397831904194447\\
6.6	0.18447700048613\\
6.7	0.117372207085086\\
6.8	0.0312392805233728\\
6.9	-0.0639468099812319\\
6.95	-0.0985914697099766\\
7	-0.16963157333273\\
7.2	-0.381105338783659\\
7.3	-0.472778644805114\\
7.45	-0.569279209041596\\
7.5	-0.586551215800714\\
7.55	-0.593066738801859\\
7.6	-0.58803405497212\\
7.65	-0.572191191602261\\
7.75	-0.511800460932449\\
7.85	-0.40560056073663\\
7.95	-0.266929263641689\\
8.05	-0.0984063646766486\\
8.15	0.0873040300050398\\
8.2	0.118711126671734\\
8.25	0.266423471943742\\
8.35	0.419574245702972\\
8.4	0.48001295172795\\
8.45	0.530220511273789\\
8.5	0.568118448780204\\
8.55	0.594511130583676\\
8.65	0.610253229920115\\
8.7	0.60183609771838\\
8.75	0.581756265546858\\
8.8	0.549353618042744\\
8.9	0.456890150502463\\
8.95	0.400546115366918\\
9	0.368470361048804\\
9.05	0.272151885471004\\
9.25	0.00529801269698282\\
9.35	-0.105806946031933\\
9.4	-0.144992241609135\\
9.45	-0.192640597471776\\
9.55	-0.258530453614718\\
9.6	-0.28050431681374\\
9.7	-0.303033496813836\\
9.8	-0.314028453155323\\
9.85	-0.313712388760369\\
10.05	-0.281407789525492\\
10.2	-0.240190694359896\\
10.25	-0.22305693767602\\
10.4	-0.135015404886797\\
10.55	-0.0271218759525773\\
10.6	0.0124162983307912\\
10.65	0.0439697934996275\\
10.7	0.0943286763021121\\
10.75	0.127719147573824\\
10.8	0.176001623376411\\
10.9	0.253999484267268\\
10.95	0.28520126121354\\
11	0.327655472754159\\
11.05	0.356146945196613\\
11.1	0.396114226654841\\
11.2	0.45465596638714\\
11.25	0.477576088101539\\
11.3	0.493030144471842\\
11.35	0.499030189070023\\
11.4	0.49516722098614\\
11.45	0.48214427776864\\
11.5	0.45807120953447\\
11.55	0.423378090780787\\
11.6	0.377813543921381\\
11.7	0.259300722813666\\
11.8	0.111297919349518\\
12.1	-0.387547255778316\\
12.15	-0.459020945833519\\
12.2	-0.520354189137443\\
12.25	-0.566229246520805\\
12.35	-0.619920828565331\\
12.4	-0.642996049163738\\
12.45	-0.643692181239157\\
12.5	-0.629692683468456\\
12.55	-0.600476994185783\\
12.6	-0.559130632695886\\
12.7	-0.461442603986455\\
13.05	0.0709187587186406\\
13.1	0.130485178257661\\
13.15	0.207475719676061\\
13.25	0.321865270142581\\
13.35	0.408284137055524\\
13.45	0.4603843125062\\
13.5	0.476952018892639\\
13.55	0.484884610889996\\
13.6	0.48621400189775\\
13.7	0.470284847667646\\
13.8	0.433188633153193\\
13.95	0.35606165968078\\
14.1	0.229493676126221\\
14.3	0.026688261701711\\
14.35	0.017465424402868\\
14.4	-0.0762752918433662\\
14.45	-0.133071527004553\\
14.55	-0.21929027201142\\
14.7	-0.318579226289572\\
14.75	-0.321165654559135\\
14.8	-0.357232818540727\\
14.9	-0.378420122107194\\
14.95	-0.382100415607781\\
15.05	-0.371804514288225\\
15.15	-0.350317325947756\\
15.3	-0.266602203086457\\
15.45	-0.162995882379956\\
15.5	-0.121436581260745\\
15.65	0.0303265980348897\\
15.9	0.287064881961044\\
15.95	0.325035163468808\\
16.05	0.42685405561215\\
16.15	0.498181300363871\\
16.2	0.522935932868098\\
16.25	0.538574786052923\\
16.3	0.543878234275066\\
16.35	0.539828108169786\\
16.4	0.522627982985853\\
16.45	0.494794072003298\\
16.5	0.456296189874504\\
16.55	0.4074123053779\\
16.65	0.286123235198755\\
16.75	0.134868128853931\\
16.8	0.105267794840682\\
16.85	-0.0333727790568084\\
17.05	-0.37072422755563\\
17.15	-0.508192005028544\\
17.25	-0.605367189472574\\
17.3	-0.636800613722002\\
17.35	-0.655548434053319\\
17.4	-0.661177087485836\\
17.45	-0.654400370961532\\
17.5	-0.637729208063259\\
17.55	-0.609640734711981\\
17.65	-0.516522696748755\\
17.75	-0.395591439608697\\
17.9	-0.167449524691282\\
17.95	-0.0909656534057675\\
18	-0.0329016682918954\\
18.05	0.00640423706030191\\
18.1	0.126595210545887\\
18.2	0.249149500346231\\
18.3	0.347978105309231\\
18.35	0.386933648713967\\
18.45	0.44457037329181\\
18.5	0.462610807500642\\
18.55	0.471829081584932\\
18.6	0.4717398837779\\
18.65	0.463782837545985\\
18.7	0.448481537426883\\
18.85	0.37309843010069\\
18.95	0.299180869590625\\
19.1	0.177198330154329\\
19.15	0.118159095320852\\
19.2	0.0702959092266404\\
19.25	0.0390691742220071\\
19.3	-0.0193281454285099\\
19.4	-0.101801473930937\\
19.5	-0.189929379945575\\
19.6	-0.263728799459521\\
19.65	-0.278717494291978\\
19.7	-0.318401764493352\\
19.75	-0.333772212386869\\
19.8	-0.357022403881242\\
19.85	-0.366313302298874\\
19.9	-0.386281015812372\\
20.1	-0.411661023625207\\
20.15	-0.414980656542372\\
};
\addplot [color=mycolor1, dashed, line width=2.0pt, forget plot]
  table[row sep=crcr]{%
0.199999999999999	-0.0159867318027409\\
0.449999999999999	-0.0121326964280115\\
1.05	0.0385415513300806\\
1.35	0.0416896291747761\\
1.75	0.0405867704970007\\
2.05	0.0560916069380184\\
2.5	0.0979863166007675\\
2.75	0.137584268414518\\
3	0.202140997653732\\
3.25	0.275889751596619\\
3.4	0.306578166540145\\
3.5	0.317643973257816\\
3.6	0.317650727771422\\
3.7	0.30538751151968\\
3.8	0.278589487935527\\
3.9	0.238182747089915\\
4	0.185571355982752\\
4.15	0.0873973253110201\\
4.3	-0.0373172431081201\\
4.5	-0.190489175533202\\
4.6	-0.252502124305295\\
4.65	-0.279894554172593\\
4.75	-0.3174813282866\\
4.85	-0.332902119972868\\
4.9	-0.332292315551989\\
5	-0.313757105215089\\
5.05	-0.298913231107651\\
5.15	-0.252783356179496\\
5.25	-0.189379067091256\\
5.4	-0.0789212573358\\
5.45	-0.0326771830970713\\
5.55	0.0459737093458834\\
5.6	0.0903108014303236\\
5.8	0.231312164255726\\
5.9	0.280366126965561\\
6	0.314164944472335\\
6.05	0.325921398565537\\
6.15	0.332266256472263\\
6.25	0.321565662677262\\
6.35	0.288774047868198\\
6.4	0.270271748134576\\
6.45	0.24539374908364\\
6.6	0.148184193255478\\
6.85	-0.0501022808914762\\
6.95	-0.134780085660097\\
7.15	-0.265584995850666\\
7.25	-0.312700733550123\\
7.3	-0.329448072683306\\
7.4	-0.344027891060851\\
7.5	-0.337229424019803\\
7.55	-0.326771821971949\\
7.65	-0.286447799722275\\
7.75	-0.227261008662978\\
8	-0.0289403565018524\\
8.15	0.100488789431633\\
8.35	0.238868964383677\\
8.45	0.282496913291673\\
8.55	0.31033309597786\\
8.65	0.318043090695621\\
8.75	0.305932917907818\\
8.8	0.295945483003269\\
8.9	0.259300513710606\\
9.05	0.184777725565205\\
9.4	-0.0423623983598134\\
9.5	-0.103408047186043\\
9.55	-0.132970077132558\\
9.7	-0.1979092388666\\
9.8	-0.226866737678574\\
9.9	-0.239499165016902\\
10	-0.23736850677545\\
10.1	-0.21701624940534\\
10.2	-0.17746804724122\\
10.3	-0.126821423426176\\
10.55	0.0502522240074477\\
10.65	0.126695276880831\\
10.85	0.261330796705796\\
11	0.321124985647963\\
11.1	0.33916987822985\\
11.2	0.336562524468437\\
11.3	0.315431615260707\\
11.4	0.270583933838687\\
11.5	0.215406566885083\\
11.8	-0.0255573251869627\\
11.9	-0.10886183502215\\
12.05	-0.210841171552808\\
12.15	-0.265593516992091\\
12.25	-0.301543085006681\\
12.35	-0.315672883592057\\
12.45	-0.308389466570034\\
12.55	-0.283618254634952\\
12.7	-0.215515568193112\\
12.8	-0.148447789644056\\
13	-0.00416045848746549\\
13.1	0.0709624706541945\\
13.15	0.11078523287792\\
13.3	0.204291346433646\\
13.45	0.26983265268229\\
13.55	0.296645908062221\\
13.65	0.307940186422172\\
13.75	0.301029255081119\\
13.85	0.277254064552807\\
13.95	0.241016190496946\\
14.05	0.190092314829787\\
14.2	0.0919789262326276\\
14.55	-0.140810685993682\\
14.7	-0.212246528385009\\
14.8	-0.241658514585023\\
14.9	-0.255693834664871\\
15	-0.250652234997748\\
15.1	-0.228046320821292\\
15.25	-0.161267285051416\\
15.4	-0.0624102508587114\\
15.8	0.213913460264127\\
15.9	0.263355807966754\\
16	0.297829854044103\\
16.1	0.314989047573039\\
16.2	0.313119101369175\\
16.25	0.30535905082554\\
16.35	0.272683774202061\\
16.4	0.251024074283222\\
16.55	0.156804813184099\\
16.75	-0.000106184523616548\\
16.95	-0.154554012987685\\
17.05	-0.218451494153783\\
17.15	-0.267168909504527\\
17.25	-0.296135496098668\\
17.35	-0.307222417172738\\
17.45	-0.298777799848828\\
17.55	-0.272220932870816\\
17.65	-0.230431865129162\\
17.75	-0.175008795602409\\
17.95	-0.0495569341575965\\
18	-0.00995244990846089\\
18.1	0.0572581805258636\\
18.2	0.116712607770353\\
18.3	0.172081794113499\\
18.4	0.211572870571768\\
18.5	0.236758848085397\\
18.6	0.247973478438325\\
18.7	0.245141006406669\\
18.75	0.23851700382772\\
18.85	0.211820140881414\\
19	0.145346927468065\\
19.1	0.0907553996407096\\
19.2	0.0336559451292047\\
19.3	-0.0268752469484426\\
19.45	-0.114415965891801\\
19.5	-0.144839317125093\\
19.95	-0.352384935273403\\
20.1	-0.400171180951777\\
20.15	-0.410375000864558\\
};
\end{axis}

\begin{axis}[%
width=\columnwidth,
height=1.5in,
at={(0.758in,0.481in)},
%scale only axis,
xmin=0,
xmax=20,
xlabel={Time (s)},
ymin=0.5,
ymax=2,
ylabel={Z (m)},
axis background/.style={fill=white},
axis x line*=bottom,
axis y line*=left,
xmajorgrids,
ymajorgrids
]
\addplot [color=black, line width=2.0pt, forget plot]
  table[row sep=crcr]{%
0.199999999999999	0.800000000000001\\
0.5	0.790977606634211\\
3.05	0.803452887800841\\
3.35	0.793879041774179\\
3.6	0.806577956754801\\
3.8	0.811559205071113\\
4.45	0.799111722572025\\
4.95	0.812619344558939\\
5.25	0.803842989875559\\
5.6	0.795188006611838\\
5.95	0.806318192948591\\
6.25	0.811796643838818\\
7	0.811637308751049\\
7.45	0.822546585947297\\
7.7	0.813563700035335\\
8.05	0.798250898451965\\
8.3	0.802045695728832\\
8.7	0.810756472731249\\
9.5	0.804282111864719\\
9.95	0.820012754990501\\
10.2	0.813720551576168\\
10.6	0.798806707817366\\
10.9	0.806528436780258\\
11.25	0.813887420946017\\
11.95	0.810541849504379\\
12.5	0.829064048734772\\
12.75	0.817305616724365\\
13.05	0.800562683914748\\
13.3	0.802752714862674\\
13.7	0.81209392661086\\
14.05	0.800775419219068\\
14.3	0.798486642170083\\
14.6	0.811567833654816\\
14.9	0.822622825517911\\
15.15	0.817621693692509\\
15.65	0.797028950501058\\
15.95	0.808847572674853\\
16.25	0.816264341653017\\
16.95	0.810050283387927\\
17.4	0.828963386289903\\
17.6	0.823244396625118\\
18.1	0.793450994918803\\
18.35	0.800744300859535\\
18.7	0.812151006238054\\
18.95	0.805040541618897\\
19.15	0.800641873203453\\
19.4	0.813629111096436\\
19.6	0.818256083172766\\
20.05	0.803645962484332\\
20.15	0.807119471871236\\
};
\addplot [color=mycolor1, line width=2.0pt, forget plot]
  table[row sep=crcr]{%
0.199999999999999	0.750987942111905\\
0.449999999999999	0.755314207470796\\
0.600000000000001	0.775868397290871\\
0.75	0.787919406429594\\
1.2	0.801159618541764\\
1.4	0.818542990325923\\
1.55	0.818795759585647\\
1.85	0.807970859297598\\
2.25	0.828973815811377\\
2.45	0.822540774779146\\
2.6	0.830949854497437\\
2.8	0.84087028387383\\
3	0.828932377645053\\
3.2	0.824085914929686\\
3.45	0.829257321894055\\
3.7	0.814382205702827\\
3.8	0.820660612033706\\
3.95	0.845403332817792\\
4.1	0.869958938825121\\
4.2	0.874339567881925\\
4.3	0.865971282540599\\
4.5	0.830050308593147\\
4.6	0.829621382722792\\
4.7	0.845718736931396\\
4.75	0.863410711709545\\
4.8	0.869133851085909\\
4.85	0.895620506754657\\
5	0.927756157957941\\
5.1	0.92303324366183\\
5.2	0.903131866278379\\
5.3	0.87529863216583\\
5.45	0.851380946742953\\
5.55	0.858524447865772\\
6	0.916164785645091\\
6.7	0.921336861445067\\
6.75	0.91970848202774\\
6.8	0.903366614807908\\
6.85	0.912037782619368\\
7.05	0.895857615108604\\
7.1	0.880719908385242\\
7.15	0.889474581891061\\
7.2	0.886081936551566\\
7.25	0.894243611968772\\
7.35	0.886413847506788\\
7.4	0.895369142992962\\
7.45	0.893915856742858\\
7.65	0.910151675584814\\
7.75	0.906538106879033\\
7.9	0.881527863850419\\
8.1	0.841583045349129\\
8.2	0.836524839526646\\
8.3	0.852846716782675\\
8.55	0.913898881392686\\
8.7	0.920522778711867\\
8.8	0.90543644826262\\
9.05	0.854054621351604\\
9.1	0.850194854350971\\
9.15	0.856895048930628\\
9.2	0.850994190813974\\
9.6	0.870572971731658\\
10.15	0.840305288440963\\
10.55	0.830266980087991\\
10.75	0.848421076970453\\
10.85	0.857331692121384\\
11.05	0.862363871014907\\
11.2	0.857650537625279\\
11.4	0.875375537708493\\
11.55	0.883023930316803\\
11.65	0.875642371740565\\
11.75	0.861097211179736\\
11.85	0.843227621697551\\
11.9	0.84307316203078\\
11.95	0.831461813523447\\
12.05	0.836889125716862\\
12.2	0.871944030240506\\
12.35	0.90634776039602\\
12.4	0.916924193264258\\
12.5	0.917367489270859\\
12.6	0.894081951680437\\
12.85	0.812142671230113\\
12.95	0.805702769056499\\
13.05	0.816705481870848\\
13.15	0.839140762900222\\
13.3	0.873286311620259\\
13.4	0.885422782271245\\
13.5	0.887937478710931\\
13.7	0.873739486531139\\
13.9	0.863190040537077\\
14.1	0.861729844722092\\
14.15	0.85043375239179\\
14.2	0.872832958294321\\
14.25	0.865788653713089\\
14.3	0.875743964402883\\
14.5	0.883337781390171\\
14.75	0.900401139872329\\
14.8	0.888691766187709\\
15.25	0.85127002463334\\
15.5	0.863508258000454\\
15.7	0.874633950524181\\
16.05	0.874411665954153\\
16.2	0.895062020003415\\
16.35	0.914838687910407\\
16.45	0.922554970957979\\
16.6	0.910783754383221\\
16.85	0.88167436202097\\
16.95	0.884274805773796\\
17.35	0.948894906098054\\
17.45	0.944802009815547\\
17.55	0.929075096397757\\
17.85	0.848938392124815\\
17.95	0.839412960643802\\
18.05	0.84295396313161\\
18.15	0.859104370138269\\
18.3	0.874021569532083\\
18.45	0.877236795740899\\
18.6	0.86501566646103\\
19.05	0.818618573840151\\
19.1	0.827444628282358\\
19.15	0.804950312232265\\
19.3	0.815513869597286\\
19.35	0.841083089897463\\
19.55	0.863411613669804\\
19.6	0.862161197528209\\
19.65	0.869049970929954\\
19.7	0.84835000221948\\
19.75	0.854518737086064\\
19.8	0.839201495242371\\
19.85	0.855893353281946\\
19.9	0.832362670923924\\
20	0.834236735753031\\
20.1	0.836678031787343\\
20.15	0.844182106103837\\
};
\addplot [color=mycolor1, dashed, line width=2.0pt, forget plot]
  table[row sep=crcr]{%
0.199999999999999	1.45608119442294\\
0.649999999999999	1.46956345010288\\
1.05	1.49588234100312\\
1.95	1.51744410560447\\
2.25	1.52442078829018\\
2.85	1.54440871231563\\
3.1	1.53916303247006\\
3.3	1.53395233467046\\
3.8	1.53912645660922\\
4.1	1.53957803987608\\
4.7	1.54890816040502\\
5.05	1.55241110945726\\
5.85	1.60559105935321\\
6.2	1.62088864911819\\
6.95	1.63435574455276\\
7.25	1.61711820655768\\
7.5	1.60443397981103\\
7.7	1.60649107761344\\
8.3	1.62355596049006\\
8.85	1.6273598172395\\
9.25	1.63160764669788\\
9.6	1.63097865155588\\
10.25	1.61386392691916\\
10.65	1.63435055436723\\
11	1.64781247275392\\
11.35	1.6450483787993\\
11.75	1.64662689004513\\
11.95	1.64111748115937\\
12.4	1.61618643571042\\
12.6	1.61829800028648\\
13.1	1.64808348920661\\
13.35	1.66547890794277\\
13.75	1.67854413603006\\
14.2	1.69282280665261\\
14.5	1.69690555492334\\
15.35	1.68868243389113\\
15.9	1.71892636195993\\
16.55	1.72672363910939\\
16.8	1.73162755659208\\
17.7	1.73024098535212\\
18.2	1.75013117291304\\
18.45	1.74468428843983\\
18.9	1.73378649174885\\
19.15	1.74044562019902\\
19.7	1.76594220531284\\
20	1.76640187359722\\
20.15	1.77922631387257\\
};
\end{axis}
\end{tikzpicture}%